\documentclass[11pt,a4paper]{article}
\usepackage[T1]{fontenc}
\usepackage{lmodern}
\usepackage[margin=0.6in]{geometry}
\usepackage{microtype}
\usepackage{amsmath,amssymb}
\usepackage{graphicx}
\usepackage{booktabs,tabularx,array}
\usepackage{enumitem}
\usepackage[numbers,sort&compress]{natbib}
\usepackage{xcolor}
\usepackage{caption}
\usepackage{placeins}
\usepackage[hidelinks,breaklinks=true]{hyperref}
\usepackage{url}
\newcolumntype{Y}{>{\raggedright\arraybackslash}X}
\setlist{nosep,leftmargin=*}

="[fonts/AstroupCompanyName.otf]" at 8pt
\DeclareRobustCommand{\astroupChinese}{{\companyChinese 向上跃升（北京）科技有限公司}}
\makeatletter
\def\@IEEENORMtitlevspace{0.5\baselineskip}
\def\@IEEEMINtitlevspace{0.25\baselineskip}
\makeatother

\hypersetup{pdfauthor={Zizhe Wang; Jia Deng; Xin Lu; Zehao Jia; Yi Huang; Yao Zhu; Jianing Li},pdftitle={Foundation Models for Autonomous Space Robotics: Architectures, Physical Adaptation, and Validation}}

\title{\textbf{Foundation Models for Autonomous Space Robotics: Architectures, Physical Adaptation, and Validation}}
\author{\normalsize Zizhe~Wang\textsuperscript{1}, Jia~Deng\textsuperscript{1}, Xin~Lu\textsuperscript{1}, Zehao~Jia\textsuperscript{1},\\[2pt]
\normalsize Yi~Huang\textsuperscript{2}, Yao~Zhu\textsuperscript{2}, and Jianing~Li\textsuperscript{3}\\[7pt]
{\normalsize \textsuperscript{1}Astroup Lab, Beijing, China}\\[2pt]
{\normalsize \textsuperscript{2}Harbin Institute of Technology, Beijing Research Institute}\\[2pt]
{\normalsize \textsuperscript{3}Peking University}\\[7pt]
{\small \{\href{mailto:wangzizhe@astroup.com}{wangzizhe},\,
\href{mailto:dengjia@astroup.com}{dengjia},\,
\href{mailto:luxin@astroup.com}{luxin},\,
\href{mailto:jiazehao@astroup.com}{jiazehao}\}@astroup.com}\\[2pt]
{\small \href{mailto:24B903024@stu.hit.edu.cn}{24B903024@stu.hit.edu.cn}\quad
\href{mailto:ee_zhuy@zju.edu.cn}{ee\_zhuy@zju.edu.cn}\quad
\href{mailto:lijianing@pku.edu.cn}{lijianing@pku.edu.cn}}}
\date{}

\begin{document}
\maketitle
\thispagestyle{plain}
\footnotetext[1]{Astroup Lab: \astroupChinese.\quad\href{https://astroup.com}{astroup.com}}
\vspace{-1.5em}
\begin{abstract}
Space robots must interpret tasks and act despite unfamiliar dynamics, difficult sensing conditions, and limited opportunities for human intervention. Foundation models offer reusable perceptual and behavioral knowledge, but their value depends on how that knowledge reaches physical execution. This critical review follows space-specific requirements through perception, prediction, planning, and action to deployment and validation, emphasizing orbital manipulation and proximity operations. Terrestrial comparisons establish benefits of particular pretraining stages and model interfaces; their transfer to space dynamics and constrained onboard operation remains insufficiently tested. Event cameras offer timely motion observations across strong illumination contrasts, while spiking neural networks provide possible routes to sparse local computation. We examine how sensor degradation, state estimation and computing choices affect feedback and recovery. Higher action throughput need not yield fresher feedback; lower processor power can also coexist with higher energy per inference. We organize validation as an iterative sim-to-ground-to-space pathway, connecting physical tests with telemetry-based diagnosis and closed-loop revalidation. A satellite-capture example links observations to approach, contact and recovery, while a corresponding protocol separates generalization from adaptation with limited target-domain interaction. These comparisons identify what models already contribute and the physical evidence needed to support autonomous space operation.
\end{abstract}

\section{Introduction and scope}
\label{sec:intro}

Space robots must interpret tasks and act with limited human intervention. Tracking or control errors can cause collisions; repair may be impossible. Satellite capture couples tracking, approach, contact, and stabilization; surface sampling connects terrain assessment to tool--material interaction; cabin maintenance requires manipulation in confined, shared workspaces. These operations rely on established dynamics, estimation, planning and supervisory autonomy~\cite{V01,V02}. Foundation models offer reusable knowledge and behavior when task-specific interaction data are scarce. Can reuse remain useful under changed dynamics, sensing and computation?

\subsection{Definitions and scope}
We define a \emph{foundation model} by broad pretraining and downstream reuse through a shared model or backbone. Size or architecture alone does not establish this status~\cite{A01,A03,A04}. We include space applications reusing identifiable foundation models and robot models with documented pretraining and reuse, assessing transfer to held-out tasks separately.

Vision--language models (VLMs) connect visual observations to linguistic or structured outputs; vision--language--action models (VLAs) extend this interface to actions. World models predict action consequences and qualify as foundation models when their pretraining and reuse support that description. Spiking neural networks (SNNs) use spikes for computation; their generality depends on training and reuse. SNNs, event cameras, task-specific reinforcement learning and conventional control inform adaptation, deployment and evaluation of pretrained components.

We emphasize orbital manipulation and proximity operations, using intravehicular and selected planetary studies to compare shared problems and physical requirements. Remote sensing and scientific question answering enter only when relevant to robot operation. We distinguish stabilized spacecraft, free-floating manipulators, pressurized cabins and lunar rovers.

\subsection{Position relative to existing surveys}
Existing surveys frame this discussion (Table~\ref{tab:survey-position}). AI-SRO connects physics, model/control interfaces, simulators, deployment resources and progressive validation~\cite{F01}. The multimodal space-robot review organizes work by task planning and end-to-end control; our comparison is limited to its original abstract~\cite{E13}. The closed loop is a shared organizing principle.

We examine how experiments distinguish reused components' contributions from task-specific adaptation and surrounding control, including timing, energy demands, new tasks and physical conditions. Within-study comparisons trace reused components, target-domain updates and tested conditions. Event cameras and SNNs receive particular attention because observation timing and sustained local computation connect model capability to physical feedback~\cite{N01,L01,L02}. We organize validation into a sim-to-ground-to-space pathway and propose a capture protocol separating held-out generalization from adaptation under a fixed interaction budget.

\begin{table}[t]
\centering
\small
\setlength{\tabcolsep}{4pt}
\caption{Existing reviews and the questions pursued here. Their emphases overlap; the final column identifies our focus in relation to each body of literature.}
\label{tab:survey-position}
\begin{tabularx}{\linewidth}{@{}>{\raggedright\arraybackslash}p{0.24\linewidth} Y Y@{}}
\toprule
Review & Established emphasis & Focus of this review \\
\midrule
Robot foundation models~\cite{A01,A02,A04} & Applications, interfaces, model use and construction & Reuse of pretrained components and evidence of transfer \\
Space robotics~\cite{V01,V02} & Autonomy, manipulation and physical requirements & Adaptation through observations, training and execution \\
AI-SRO~\cite{F01} & Capability lifecycle: physics, interfaces, resources and validation & Gains, failures and tested conditions; timing and energy demands \\
Multimodal space robots~\cite{E13} & Task planning and end-to-end control (abstract inspected) & Joint tests of task transfer and physical variation \\
Neuromorphic methods~\cite{N01,N02,L01,L02,L03} & Sensing, representations, training and hardware & Roles of event sensors and spiking computation; deployment cost \\
\bottomrule
\end{tabularx}
\end{table}

\subsection{Literature selection}
This critical narrative review considers works available by 14 October 2026. Initial searches were followed by targeted retrieval and source checks on 15 October, retaining the original eligibility cutoff. Surveys seeded citation tracing, and web searches led to primary sources through arXiv, publishers, proceedings, institutions, and space agencies. Selection favored distinct operational interfaces, physical mechanisms, or validation settings; terrestrial studies explain transferable methods. Titles, DOIs, and versions supported deduplication.

The ancillary inventory records the search history, versions, and reading scope of all cited works. Mechanism comparisons use methodological passages; abstract-only sources support only their stated claims. Unresolved candidates and a local unpublished survey were excluded. Purposive selection limits completeness, and differences between tasks preclude pooled performance rankings.

The review proceeds from space requirements through the embodied closed loop to adaptation, validation and cross-study conclusions.


\section{From space operating conditions to technical requirements}
\label{sec:requirements}

\begin{figure}[t]
\centering
\includegraphics[width=\linewidth]{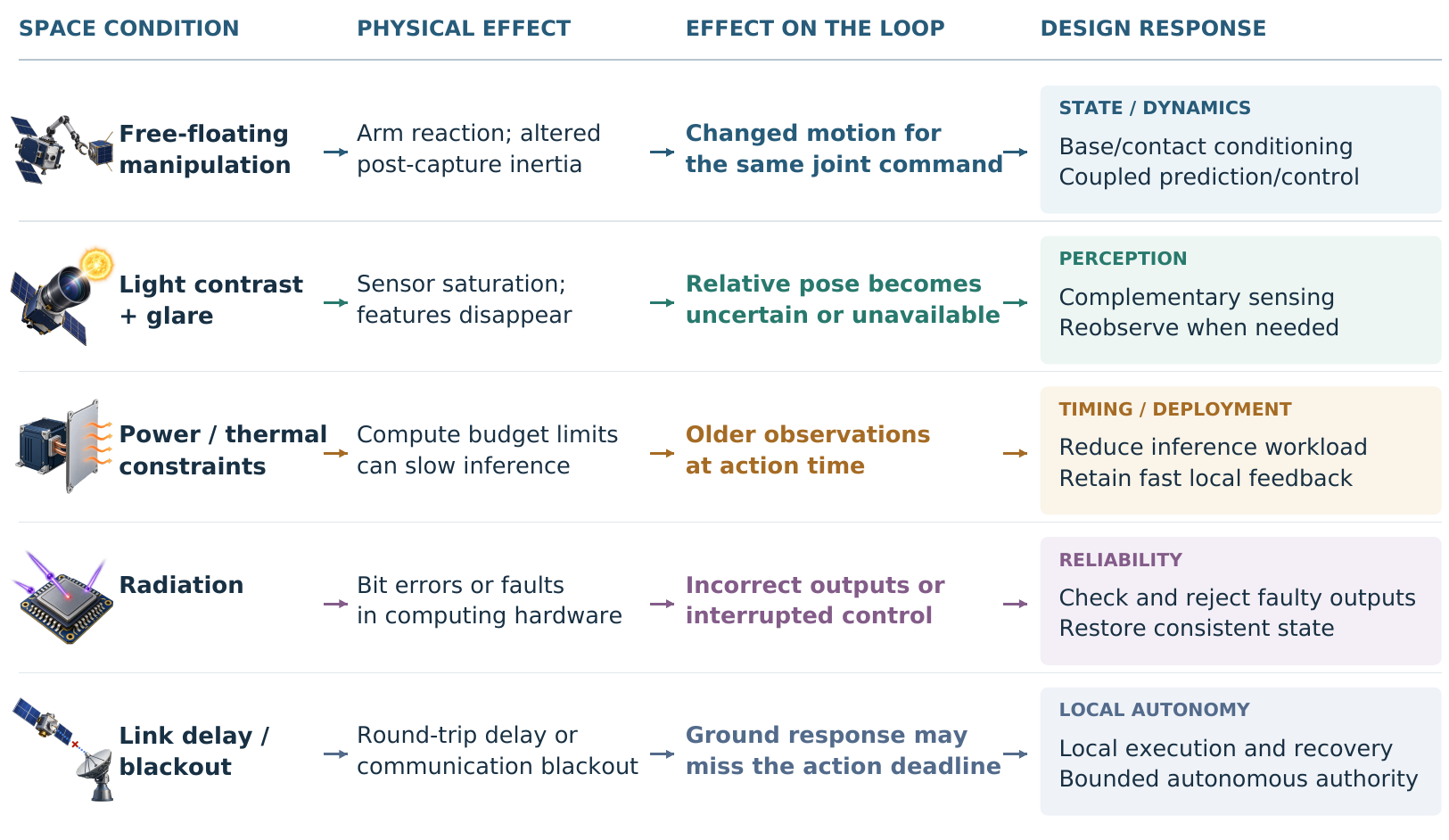}
\caption{How space conditions change the autonomy loop. Each pathway links a physical or operational constraint to its consequences for execution and to model or control responses. The pathways illustrate mechanisms discussed in this review; their importance depends on the mission and hardware~\cite{V01,V02,N01,V19,V20}.}
\label{fig:matrix}
\end{figure}

Space deployment changes sensing, prediction and control. Figure~\ref{fig:matrix} traces how space conditions change action consequences, observation quality, feedback timing and recovery, motivating corresponding model and control choices.

\subsection{Dynamics and contact}
\begin{figure}[!htbp]
\centering
\includegraphics[width=\linewidth]{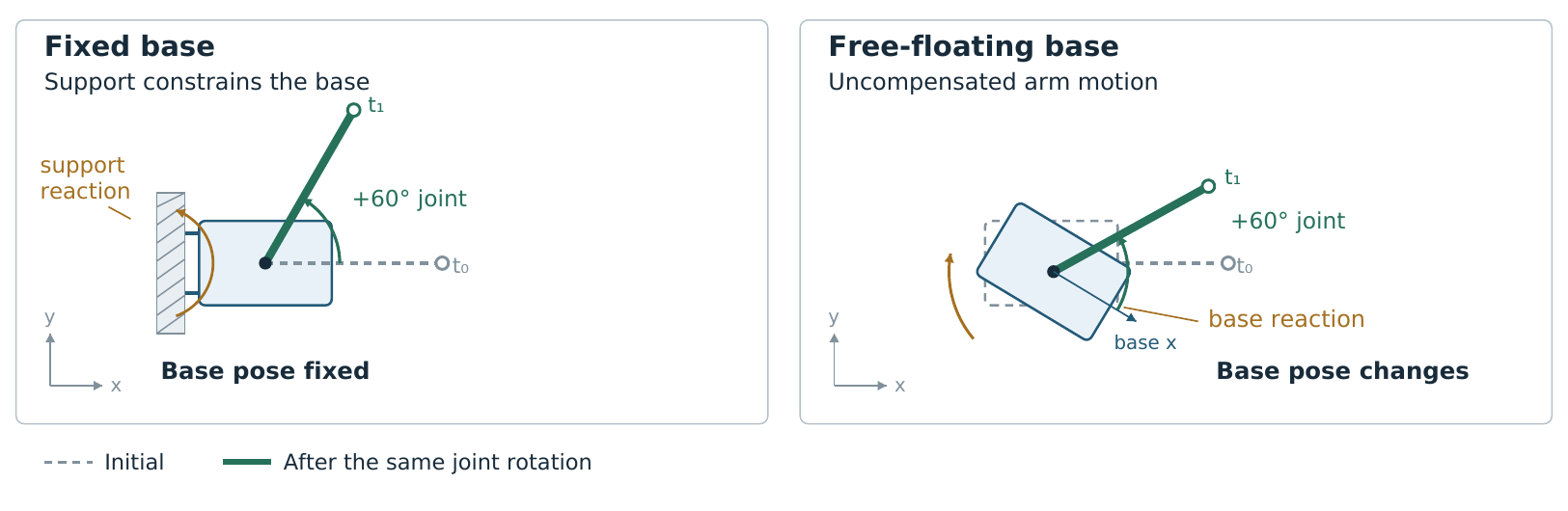}
\caption{The same joint rotation with a fixed or freely moving base. Dashed and solid geometry denote initial and final poses. Support reactions hold the fixed base; the free base translates and counter-rotates. This planar example starts at rest in a local free-fall frame, neglecting external disturbances and active base control during a short maneuver. The response depends on mass distribution and geometry; a learned policy must account for the resulting end-effector motion~\cite{P07,V01}.}
\label{fig:dynamics-primer}
\end{figure}

A fixed-base terrestrial arm can often treat its base frame as stationary. A free-floating arm exchanges momentum with the base, changing the base attitude, the robot's reachable configurations and the required control effort. Figure~\ref{fig:dynamics-primer} illustrates why the same joint motion can produce a different end-effector motion. Capture changes mass distribution and motion. ETS-VII experiments and the manipulation literature establish these effects independently of learned policies~\cite{P07,V01}.

\begin{samepage}
Rendezvous requires relative states and orbital dynamics~\cite{E05,V13}; capture adds base--target coupling and contact. State and dynamics models must follow the mission stage.\par
\end{samepage}

For orbital manipulation, relevant state includes joint configurations, base attitude and velocity, target-relative motion, and contact mode. Active attitude control also introduces actuator and propellant constraints. Task-specific space reinforcement learning illustrates these state requirements~\cite{P01,P02}. Surface interaction poses a different problem: reduced gravity changes both robot loading and granular-medium behavior. Parabolic-flight wheel--soil experiments found reduced traction and increased sinkage relative to ground tests at matched wheel loads~\cite{U08}. Reducing the robot's supported weight therefore does not reproduce all relevant soil effects.

\subsection{Perception and hardware constraints}
Illumination extremes and specular reflections can obscure spacecraft features, as the physical images in SPEED+ illustrate~\cite{V12}. Thermal changes also perturb measurement geometry: MESSENGER images reveal temperature-dependent focal length~\cite{C03}. Proton irradiation of CMOS imagers produces increased dark current and switching pixel defects~\cite{C02}. Missing or displaced keypoints can then bias pose estimates or interrupt tracking. The resulting requirement is to expose calibration state and measurement validity to the estimator and learned policy, alongside coordinate frames and timestamps.

Temperature-dependent calibration, dark-signal correction, and bad-pixel masks address characterized errors; learned compensation should be evaluated by pose and tracking accuracy. Neither calibration nor a plausible reconstruction restores a unique measurement where saturation or occlusion has removed it. The robot may need a new exposure, viewpoint, or modality. Event sensing improves temporal sampling~\cite{N01}, while force, touch, and proprioception supply contact or body-state information~\cite{G02,G03,G05}; their value depends on the state uncertainty and feedback deadline.

Sustained computation depends on power and heat rejection. External vacuum removes ambient convective cooling, making structural conduction and radiation central; pressurized cabins and planetary atmospheres differ. Mounting, conductive paths, duty cycle, and thermal-control capacity determine feasible computation~\cite{V19}. Model deployment must fit the spacecraft's peak and sustained power, allowable temperature, and task-energy budgets.

Radiation can disrupt computation and storage through single-event effects and accumulated dose~\cite{V20}, requiring error detection, redundancy, recovery, and appropriate devices. Proton and neutron tests measure device responses beyond what image-domain adaptation can establish~\cite{P10,P11}. Redundancy also consumes power and may remain vulnerable to common failures, so protection and computational efficiency must be assessed together.

\subsection{Communication and limited interaction data}
Communication delay, jitter, and outages limit reliance on human intervention. SUPVIS Justin demonstrated task-level supervision by ISS astronauts while a ground robot planned and executed locally~\cite{U10}. Collision avoidance and contact response need local feedback; Section~\ref{sec:planning} examines the division of authority.

Hardware access and the consequences of failure restrict interaction data. Terrestrial pretraining supplies object semantics and action structure, but representative floating-base contacts and recovery trajectories remain difficult to acquire. Studies should identify physical variations absent from pretraining and the new interaction needed to handle them~\cite{A01,N25}.

\subsection{Connecting requirements to a task}
Consider satellite capture. Visual pretraining may identify a grasp feature, but its location at contact depends on target-relative motion and observation age. Approach needs pose tracking and uncertainty propagated to the intended action time~\cite{V13}; closing the gripper adds contact evidence, and stabilization requires the coupled base--target state~\cite{V01}. Degraded or stale observations can therefore require a slower approach or reacquisition before engagement. Adaptation data and performance across targets assess reuse; contact loads, base disturbance, and stabilization assess execution. This task connects the following architecture to the evaluation in Section~\ref{sec:validation}.


\section{An embodied closed loop for model integration}
\label{sec:architecture}

\subsection{Functional interfaces and feedback}
\begin{figure}[t]
\centering
\includegraphics[width=\linewidth]{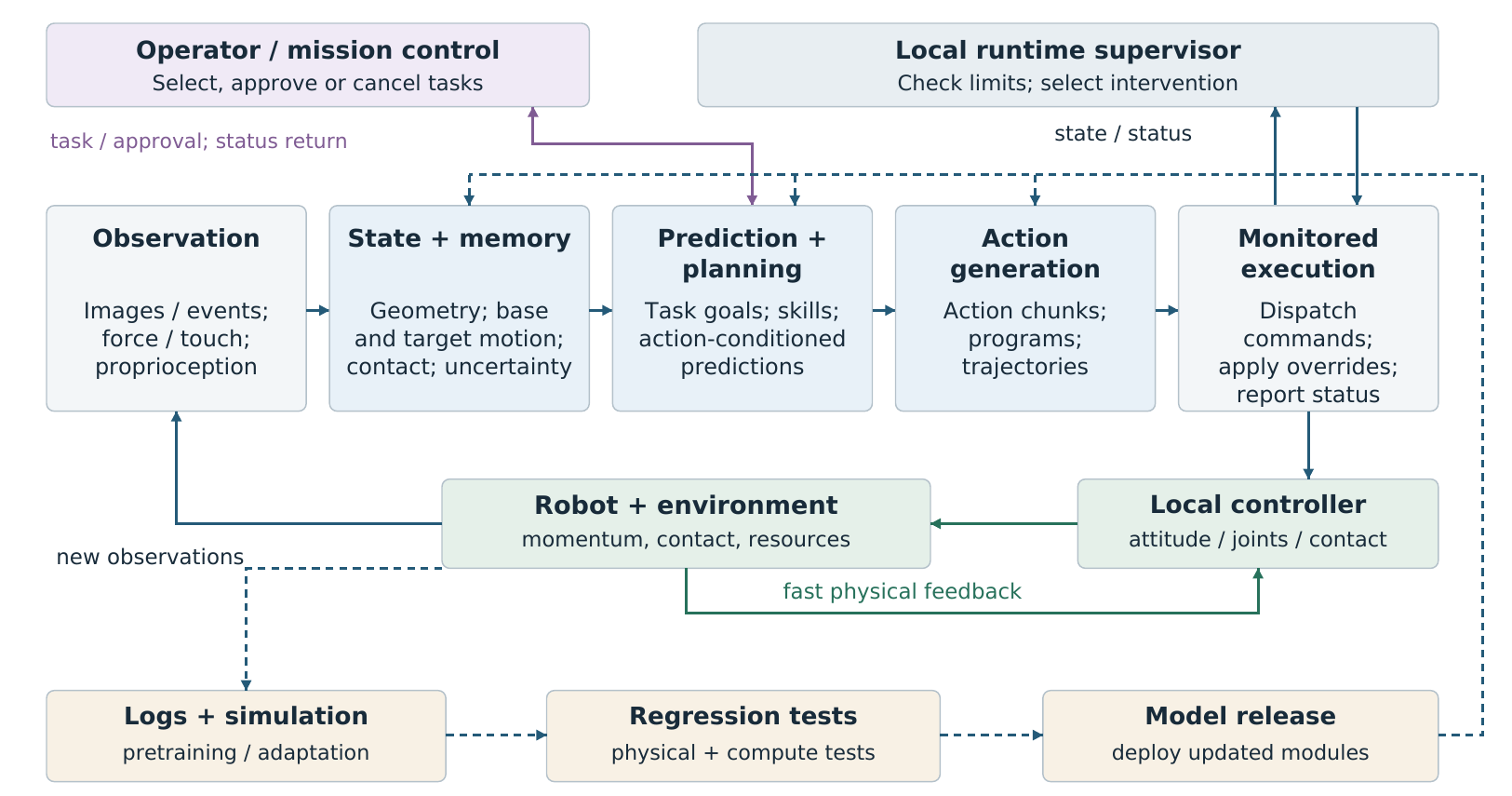}
\caption{A functional architecture for embodied space autonomy. Operators propose and approve tasks; local runtime supervision monitors execution, and status returns to mission control. Fast physical feedback remains local. Dashed arrows show data collection, offline adaptation, and component release. Blue boxes indicate possible pretrained modules; foundation models may span several functions~\cite{F14,F18,F19,A05,A08,U10}.}
\label{fig:loop}
\end{figure}

Embodied systems act on the environment and revise decisions from subsequent observations. Figure~\ref{fig:loop} organizes this loop into observation, state and memory, prediction and task planning, action generation, and monitored execution. A unified VLA can span several of these functions, while a modular system can combine pretrained models with conventional estimators and controllers. Inputs, outputs, update rates and feedback paths reveal each model's contribution~\cite{A02}.

\emph{Grasp the handle} requires locating it, estimating relative motion and detecting contact. The planner needs feedback distinguishing attempted and completed actions. Inner Monologue shows how such feedback can improve recovery, while errors in the success detector can mislead planning~\cite{A09}. Interfaces should therefore carry observable task status together with timestamps, coordinate frames, and indications of invalid estimates or failed actions.

Foundation models can produce annotations, synthetic training data, or distillation targets offline. Models trained on these outputs need separate evaluation of task performance and resource use.

\subsection{How models are coupled to execution}
Model integration depends on its outputs, training and execution feedback. A language model may select a complete skill, generate a program, or issue an incremental motion command. A robot policy instead learns an observation-to-action mapping from interaction data. Predictive models can support either approach by supplying training experience or evaluating candidate actions online. These are combinable design choices. Table~\ref{tab:model-interfaces} compares their physical assumptions and feedback responsibilities.

Explicit interfaces expose skill identifiers, programs, motion targets, or trajectory constraints. SayCan scores skills using semantic relevance and feasibility; Code as Policies generates programs that call perception and control routines~\cite{F18,A08}. Selecting a whole skill delegates its feedback loop, whereas issuing successive motion commands makes model latency part of that loop. Typed choices restrict the available commands, while the executor determines their physical meaning~\cite{T09}. A learned warm start supplies another interface: it initializes an optimizer that retains explicit dynamics and constraints~\cite{A05}.

\begin{samepage}
Learned robot policies include OpenVLA and $\pi_0$, which inherit visual--language components, and Octo, whose backbone is pretrained across robot datasets~\cite{F14,F15,F16}. Their decoders learn action conventions and observation fusion; fast feedback may still come from a lower-level controller. Calling a robot API alone does not establish this training history or make a language model a VLA.\par
\end{samepage}

Prediction can support either interface: UniSim supplies interactive training experience, while Cosmos Policy can use predicted outcomes and values to select candidates online~\cite{G10,G13}. Table~\ref{tab:model-interfaces} locates physical state and feedback within these routes; Figure~\ref{fig:capture-feedback} follows their implications through a capture attempt.

\subsection{Timing, supervision, and model updates}
Task reasoning, action generation, and servo control operate at different rates. A policy may execute only part of a predicted sequence before processing a new observation, while a faster controller supplies feedback during execution. Prediction horizon, execution interval, inference delay, and observation age therefore describe different aspects of responsiveness; Section~\ref{sec:action} defines them.

The local supervisor can reject or interrupt actions. Abate et al. demonstrate ground intervention using a backup controller to maintain a line-of-sight constraint~\cite{U07}. Humans select and parameterize tasks~\cite{U10}; conversational advice remains separate from control~\cite{E01}. The architecture must specify approval and cancellation authority, status feedback, and permitted local behavior when assistance misses a task deadline.

Updates to model weights or skills alter future behavior and require separate evaluation from recovery with existing capabilities. Operational logs can support adaptation, followed by regression tests before the updated component is released.


\section{Multimodal perception and state representation}
\label{sec:perception}

Planning and control require estimates of target-relative motion, spacecraft-base motion, joint states, and contact~\cite{V01}. We collect these quantities in a task state, $x_t=(x_t^{\mathrm{rel}},x_t^{\mathrm{base}},q_t,\dot q_t,c_t)$. Pretrained semantic and geometric features must connect to these physical quantities.

\subsection{Connecting semantic features to physical state}

Space-LLaVA adapts visual--language knowledge to extraterrestrial imagery and evaluates it on offline image--language tasks~\cite{F02}; the same paper uses GPT-4o in a separate lunar-rover simulation. To guide motion, semantic features must also support metric estimates. A planetary visual-odometry pipeline uses SAM rock landmarks and DINOv2 descriptors before geometric motion estimation, obtaining metric scale from simulator depth on LuSNAR or stereo disparity on Katwijk analogue sequences~\cite{F10}.

FoundationPose estimates object pose through render-and-compare refinement and ranking, using synthetic pretraining and RGB-D observations~\cite{U09}. It requires a CAD model or reference views with provided or estimated poses; the latter are used to fit an object-specific neural field. Novel-object transfer thus depends on geometric information; noncooperative targets may still have known geometry.

A specialist pipeline feeds SPNv2 image measurements into an unscented Kalman filter to track relative orbital state, attitude, and angular velocity~\cite{V13}. It propagates state between measurements and adapts process-noise covariance; SHIRT evaluates it on synthetic and ground-testbed sequences with a known target. For a capture policy, this suggests an interface containing the predicted state, uncertainty, and observation age, so that missed visual updates remain visible to action selection.

PaLM-E embeds images and continuous states into language-token space to generate textual instructions that invoke downstream policies~\cite{F19}. HPT uses visual and proprioceptive stems, a shared Transformer trunk, and action heads; transfer initializes new stems and heads while freezing the trunk~\cite{G05}. These methods connect state to language decisions or adapt interfaces around a shared representation.

\subsection{Event, spike, and single-photon sensing}

During spacecraft approach, motion and illumination contrasts can degrade features just when fresh pose updates are needed. Differential event cameras provide high dynamic range and asynchronous intensity-change measurements~\cite{N05,N07}. Combining them with image--language priors requires preserving timestamps through encoding and estimation. Asynchronous readout is distinct from radiation qualification: neutron irradiation produced noise bursts in a tested event sensor, although motion information remained usable in the reported experiment~\cite{C04}. Its radiation response and tracking benefits require separate evidence.

\begin{figure}[!htbp]
\centering
\includegraphics[width=\linewidth]{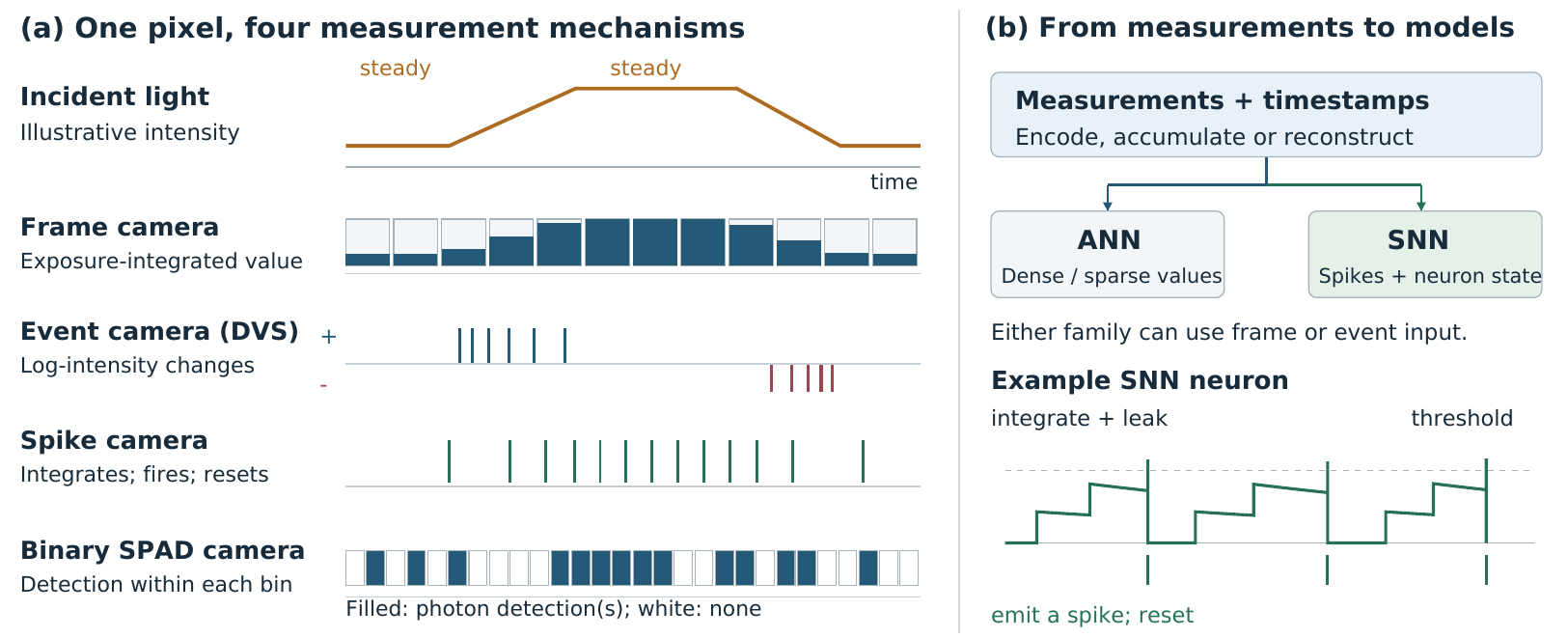}
\caption{From light to measurements and neural computation. Left: idealized single-pixel outputs share a nonquantitative time axis. Frame-bar height represents integrated light; differential events mark log-intensity changes; integrating pixels emit spikes at an accumulated-light threshold. The passive SPAD512 example records detection within each exposure; other SPAD readouts differ. Right: suitable encodings connect measurements to either network family; a leaky integrate-and-fire neuron illustrates state accumulation and spike emission. Traces are explanatory illustrations~\cite{N01,L03,T01,L02}.}
\label{fig:sensing-primer}
\end{figure}

Figure~\ref{fig:sensing-primer} compares frame, differential-event, integrating-spike and single-photon avalanche diode (SPAD) outputs under changing illumination. Table~\ref{tab:sensor-inputs} relates these measurements to model inputs and information limits. Event streams can feed conventional networks, while SNNs can process frame-camera inputs~\cite{L02}. Section~\ref{sec:adaptation} examines spiking computation.

\begin{table}[!hbp]
\centering
\small
\setlength{\tabcolsep}{4pt}
\caption{Four sensing mechanisms, input representations and information limits. Sensor outputs do not determine the neural-network implementation.}
\label{tab:sensor-inputs}
\begin{tabularx}{\linewidth}{@{}>{\raggedright\arraybackslash}p{0.22\linewidth}YY@{}}
\toprule
Sensor & Measurement & Model input and information limit \\
\midrule
Conventional frame camera~\cite{N01,T01} & Light integrated over an exposure, then read as intensity values & Image tensors; motion during exposure can blur features; saturation loses intensity detail \\
Differential event camera (DVS)~\cite{N01} & Thresholded log-intensity changes with location, timestamp and polarity & Event tensors/tokens or reconstructed images; unchanged intensity is not directly measured \\
Intensity-integrating spike camera~\cite{L03,T03} & Intensity accumulated to a threshold; each spike resets the integrator & Temporal spike encoding or reconstruction; intensity requires spike history \\
Passive SPAD camera (SPAD512)~\cite{T01,T02} & Binary value per pixel/exposure: no detection or at least one & Binary-frame accumulation estimates intensity; one exposure does not resolve photon multiplicity \\
\bottomrule
\end{tabularx}
\end{table}

Accumulating differential events into frames allows reuse of image encoders, at the cost of temporal compression over an observation window~\cite{L01}. EventBind instead adapts CLIP through temporal event encoding and prompts, aligning event, image, and text embeddings contrastively~\cite{N10}. EventGPT combines spatiotemporal aggregation with event--language adaptation to generate language responses~\cite{N09}.

E-VLA converts aligned event windows into frame-like inputs, using either an RGB overlay or adapters inside a frozen visual encoder; the resulting tokens condition a SmolVLA action expert~\cite{N11}. Event-VLA instead uses action queries and gated cross-attention to select event tokens while retaining RGB--language priors~\cite{N12}. Both are evaluated in terrestrial manipulation under degraded visibility. Spacecraft evidence includes the SPADES event-camera dataset~\cite{N05} and S2E2, a compact SNN that estimates pose using a custom split of real ground-testbed recordings from the SEENIC dataset~\cite{N06}.

For intensity-integrating cameras, SSDEFormer encodes temporal spike accumulations and uses a Transformer for stereo depth estimation, evaluated on synthetic data and real PKU-Spike-Stereo recordings~\cite{T03}. This specialist terrestrial method offers a candidate geometric input for robot policies; space-robot control remains untested.

In lunar-analogue experiments, Fast Vision in the Dark compares SPAD512 and conventional-camera imagery from a Leo Rover at LunaLab~\cite{T01}. Pretrained SAM ViT-H is tested on one selected scene under several lighting conditions; ORB-SLAM3 is evaluated on recorded rover trajectories. The teleoperated recordings come from SPICE-HL3, which also provides conventional visual and motion measurements~\cite{T02}. Integrated photon frames permit reuse of image models, but lower bit depths can fragment segmentation, and processing raw binary streams introduces bandwidth and representation challenges~\cite{T01,T02}.

\subsection{Force, touch, and proprioceptive encoding}

Force--torque sensors measure interaction wrench, while visuotactile sensors encode local deformation through images. AnyTouch learns representations across tactile sensors through shared tokenization, masked modeling, multimodal alignment, and cross-sensor matching~\cite{G03}. FuSe addresses a complementary problem: a pretrained policy may ignore a newly added sensor. It supplements imitation learning with contrastive and language-generation objectives to connect heterogeneous observations to task semantics~\cite{G01}. Together, these methods distinguish making sensor representations compatible from ensuring that a policy uses the information they contain.

For a floating robot, interpreting wrench also requires the sensor frame, calibration, joint motion, and base response. These quantities must be available to the policy or to the controller that processes force measurements. Section~\ref{sec:action} follows this information into contact-sensitive execution.

Perception supplies semantic descriptions, geometric estimates, and contact or body state. Their calibration, validity, and timing determine how policies can use them. Semantic confidence cannot substitute for pose uncertainty.


\section{Prediction and task planning}
\label{sec:planning}

Prediction estimates action consequences, task planning selects capabilities and their order, and motion planning computes feasible trajectories. Physical constraints enter through learned predictions, structured task descriptions, or the planner that turns them into motion.

\subsection{World models: prediction targets and control roles}

\begin{samepage}
Predictive models generate training experience, support action learning, or evaluate candidate actions during execution. DreamerV3 learns latent dynamics and trains a policy on imagined trajectories; its generality reflects a shared algorithmic configuration across tasks rather than a reusable pretrained model~\cite{F20}.\par
\end{samepage}

UniSim learns action-conditioned future observations through video diffusion and uses simulated interaction for policy learning~\cite{G10}. DreamGen fine-tunes an image-to-video model to generate robot videos from an initial frame and instruction, then recovers pseudo-actions for a separate policy~\cite{G12}. Learning can inherit transition or action-inference errors.

V-JEPA~2-AC predicts future representations rather than images~\cite{T05}. It freezes a pretrained video encoder and learns an action-conditioned predictor from robot trajectories. A receding-horizon planner selects actions toward a goal representation, executes the first, and replans. Franka pick-and-place experiments supply two intermediate goal images before the final goal and switch between them on a fixed schedule. The reported single-RTX-4090 configuration takes 16 seconds to compute an action. Camera sensitivity, externally supplied subgoals and planning time shape demonstrated autonomy. Space applications additionally need representations preserving relative motion and contact information.

GR-2 learns autoregressive video prediction before jointly predicting images and action trajectories with a conditional variational action decoder~\cite{G11}. Cosmos Policy adapts a video model to predict actions, future states, and values~\cite{G13}. Its direct-policy evaluation discards state and value predictions, whereas its planning mode assesses candidates with them. With equal numbers of gradient updates, pretrained initialization outperforms random initialization (Table~\ref{tab:representative-evidence}).

A specialist continual-adaptation method updates learned dynamics after robot failures~\cite{F07}. It freezes DreamerV3's observation encoding, decoding, and reward prediction while updating recurrent and latent-transition components; the actor--critic learns from the revised imagined trajectories. In simulations of wheel failures, thruster failures during microgravity navigation, and screwdriver misalignment, recovery is partial and not always sustained. Capture additionally requires representations of base motion and changing contact modes accurate enough to predict impulses and constraint violations~\cite{V01}.

\subsection{Grounding language in available capabilities}

SayCan scores skills, PaLM-E conditions language reasoning on observations, and Code as Policies calls perception and action APIs~\cite{F18,F19,A08}. Inner Monologue uses execution feedback to revise plans~\cite{A09}. VoxPoser composes spatial affordance and avoidance maps for a motion planner~\cite{T06}. It uses known robot dynamics and mostly assumes static scenes, replanning end-effector paths from observations. Its reusable spatial objective interface would need a planner predicting target motion and base--arm coupling while enforcing whole-body constraints for orbital capture.

LLM-KGFP organizes lunar operational knowledge in a graph and revises plans through feedback in lunar south-pole simulation~\cite{F06}. SpaceMind combines a VLM, skill instructions, tool calls, and execution memory~\cite{F03}. Its skill evolution revises task-specific failure-recovery instructions rather than motor weights; it shows no cross-task skill transfer or benefit where baselines already succeed. Evaluation comprises 174 UE5 runs and 18 planar ground-robot runs, without orbital dynamics. Rendezvous succeeded in 6/6 ground trials and search and approach in 4/6; one of six inspection runs collided. The collision motivates continued monitoring after command acceptance: intended motion or mechanical engagement must occur and constraints remain satisfied.

\subsection{From mission specifications to feasible motion}

OrbitTAMP maps language to orbital motion~\cite{E05}. A language model parses instructions into a partial \emph{MissionConfig}. Behavior-duration planning completes the behavior sequence, waypoint planning supplies missing relative-state coordinates, and trajectory optimization computes state and control trajectories. Constraints can include flight time, $\Delta v$ budget, and keep-out-zone dimensions. Physics enters through the planning representation and optimizer.

With downstream planning fixed, held-out tests compare language-model parsers against an ontology-aware rule parser (Table~\ref{tab:representative-evidence}). Their relative performance changes with the instruction style. Separately, two verifier-guided revisions raise Qwen3.5-9B's exact recovery from 75\% to 88\% on 100 development commands; five revisions add no benefit. Structurally valid semantic errors can escape verification, and the version-1 trajectories are nominal and open-loop. A converged optimizer solves the encoded problem, which may still differ from the intended mission.

SELENE specifies a complementary interface through Atomic Action Contracts, connecting semantic planning to VLA, RL, or optimization-based executors~\cite{F09}. Preconditions, goals, safety envelopes, and validation metadata define when a skill may run and what it must achieve. Tests cover ground LunarBot manipulation, simulated lunar-gravity traversal, and offline skill instantiation and registration. They do not yet demonstrate detecting a missing capability at runtime, creating a new skill, and executing it. Whereas OrbitTAMP completes a numerical mission specification, SELENE describes how existing capabilities are selected and called.

\subsection{Autonomous decisions and requests for assistance}

Planning also allocates human supervision. In SUPVIS Justin, ISS astronauts selected task commands and parameters for Earth-based Rollin' Justin, which planned and executed them locally~\cite{U10}. NealAI assists astronauts supervising a ground robotic team~\cite{E01}. NASA's Perseverance account describes another allocation: Claude-assisted planning on Earth, digital-twin checking before uplink, and execution on Mars~\cite{F12}. By comparison, AEGIS selects scientific targets onboard through a specialized system~\cite{A06}.

KnowNo uses held-out conformal calibration to form sets of candidate next steps; non-singleton sets trigger assistance~\cite{A07}. Its guarantee applies to semantic choices under the calibrated scenario distribution and help mechanism. Physical uncertainty and execution deviations require responses such as renewed observation or controller-level recovery, including permitted local action during communication delays. These extensions require testing and intervention accounting beyond KnowNo's semantic guarantee.


\section{Action generation and physical execution}
\label{sec:action}

Joint displacements, end-effector targets, throttle vectors, and skill identifiers have different physical meanings despite sharing numerical encodings. Transfer requires matching the action representation to the robot's controller.

\subsection{Direct commands and learned action policies}

Language models can issue motion commands through structured interfaces. GPT-3.5 and LLaMA are fine-tuned on telemetry to return throttle-vector function calls in KSPDG spacecraft pursuit--evasion simulation~\cite{E09}. With fixed weights, GPT-6-Astra instead uses images and joint feedback to command XLeRobot and generate local feedback programs~\cite{T07}. Machine-readable geometry and synchronized experience reduce completion time; the twelve physical trials use operator-confirmed button contact as their endpoint.

A broader Astra study compares direct end-effector commands with a hybrid that accepts or corrects $\pi_{0.5}$ proposals~\cite{T08}. On 50 paired RoboDojo simulation instances, sampled mainly from tasks with low published $\pi_{0.5}$ success, hybrid control completes 24 versus 13 for direct control; 85.6\% of its execution steps come from the learned policy. Here $\pi_{0.5}$ is task-finetuned; on RoboLab, where Direct performs better, the hybrid uses a DROID-trained checkpoint zero-shot. The ranking therefore reflects both task requirements and the fit of the reused policy.

Typed decision models offer a different interface. In a planar-arm simulation, Jev selects discrete torque levels or evaluates code-generated skill outcomes~\cite{T09}. Raw torque selection completes 0/20 episodes on the native task, compared with 9/20 using skills and external simulation. The comparison changes both action abstraction and predictive support: it does not show that Jev itself has learned a dynamics model. Command granularity and the source of physical prediction therefore matter alongside model identity.

Robot policies learn action mappings directly from trajectories. RT-2 predicts action tokens~\cite{F13}; OpenVLA combines DINOv2 and SigLIP features with Llama~2 and discretizes each action dimension into 256 bins spanning training-data quantiles~\cite{F14}. Scaling and command semantics determine how those tokens move the robot.

Diffusion Policy learns conditional denoising of continuous action sequences~\cite{F17}. Octo places a small diffusion head after a pretrained Transformer, so repeated denoising need not rerun the backbone~\cite{F15}. In $\pi_0$, PaliGemma initializes the visual--language component; a separate action expert processes robot states and noisy actions, generating a chunk by integrating a learned conditional vector field~\cite{F16}. Different training targets and repeated computations affect both latency and energy use.

Table~\ref{tab:model-interfaces} compares where these routes obtain physical behavior and how execution returns feedback. A shared model can support several interfaces; their suitability for orbital capture depends on the state and constraints carried through each one.

\begin{table}[t]
\centering
\small
\setlength{\tabcolsep}{4pt}
\caption{Physical assumptions and feedback responsibilities across model interfaces. Orbital adaptations are proposed extensions of the methods discussed in Sections~\ref{sec:planning}--\ref{sec:action}; prediction can support any action interface.}
\label{tab:model-interfaces}
\begin{tabularx}{\linewidth}{@{}>{\raggedright\arraybackslash}p{0.22\linewidth}YYY@{}}
\toprule
Output and examples & Source of physical behavior & Feedback responsibility & Orbital adaptation \\
\midrule
Numerical targets: Astra~\cite{T07,T08} & Robot API and servo; model uses observed state & Local servo/program; observations return to model & Base-aware frames; coupled limits; observation age \\
Skill or program calls; typed choices: SayCan, Code as Policies, Jev~\cite{F18,A08,T09} & Existing skills/controllers; external simulation in Jev skill tests & Executor closes local loop and returns status & Coupled-dynamics skills; checked preconditions and recovery \\
Spatial objectives or constraints: VoxPoser, OrbitTAMP~\cite{T06,E05} & Explicit robot/orbital dynamics in planner & VoxPoser replans; cited OrbitTAMP trajectories are open-loop & Target prediction; whole-body constraints; execution feedback \\
Learned action policies: OpenVLA, Octo, $\pi_0$~\cite{F14,F15,F16} & Robot trajectories and lower-level control & Policy updates; lower-level control between updates & Base/contact state; adapted training; constrained execution \\
Latent/video futures: V-JEPA~2-AC, Cosmos Policy~\cite{T05,G13} & Transitions learned from robot trajectories & Planner/policy selects; observations update prediction & Momentum/contact representation; calibration under changed dynamics \\
\bottomrule
\end{tabularx}
\end{table}

\subsection{Action chunks and feedback timing}

\begin{samepage}
Let $A_t^H=(u_t,\ldots,u_{t+H-1})$ contain $H$ predicted steps, of which $K\leq H$ are executed before the generating policy incorporates a new observation. Faster decoders or controllers may receive feedback within that interval. If an observation is acquired at $t_o$ and its first resulting command issued at $t_c$, its age is already $t_c-t_o$; actuator response adds further delay. Commands issued through model services incur request and response delay. Astra's local programs execute feedback loops between model calls; Jev's torque-step duration specifies simulated motion time~\cite{T07,T09}.\par
\end{samepage}

These timescales matter independently of action encoding. Cosmos Policy's direct ALOHA evaluation has a 0.95-second blocking inference pause on one H100, followed by 50 executed steps at 25\,Hz, or two seconds of execution~\cite{G13}. FAST compresses action sequences into tokens but does not determine the feedback interval~\cite{G08}. Shorter execution chunks help only when new actions arrive in time.

Real-time chunking (RTC) overlaps inference with execution, preserves the actions committed during inference delay, and inpaints the remaining chunk with softly weighted overlap~\cite{U04}. Its Kinetix ablation favors soft over hard masking particularly at lower delays; robot tests use terrestrial bimanual manipulation. RTC improves continuity between chunks, whereas SpikingVLA overlaps computation inside a network~\cite{U03}. Physical execution additionally needs an interrupt when contact assumptions fail.

\subsection{Incorporating space dynamics}

Free-floating arm motion reacts on the spacecraft base, as demonstrated by ETS-VII's generalized-Jacobian and reaction-related control experiments~\cite{P07}. Reproducing a terrestrial end-effector trajectory therefore changes base motion and control-resource use. Capture adds the target to the coupled system: contact can dissipate kinetic energy, but internal forces cannot remove its total momentum~\cite{V01}. Retaining a grasp and stabilizing the combined motion consequently require more than reaching the predicted contact pose.

Adaptation can enter at three locations~\cite{V01}. \emph{State conditioning} gives the model relative motion, base attitude and velocity, and joint state. \emph{Training adaptation} exposes it to changes in inertia, momentum, and support conditions. \emph{External control} interprets or modifies proposed targets using coupled dynamics and constraints. For explicit targets and skills, physical adaptation can modify state descriptions and executors; learned action mappings and transition models must also be checked against the new dynamics.

SpaceOctopus combines multi-arm policies with a lower-level PD controller in MuJoCo~\cite{P01}. Prior-policy-guided learning uses inverse-kinematics knowledge for dual-arm free-floating motion planning while varying base mass and target rotation~\cite{P02}. These specialist systems show where relevant state and dynamics enter the policy--controller combination. The latter stops short of post-contact force control, so extending its approach to stabilized capture requires a further contact-control stage.

\subsection{Contact sensing, action updates, and compliant control}

Contact information can influence execution at several levels. ForceVLA conditions $\pi_0$ action decoding through a force-aware mixture of experts, using tool-center-point pose, gripper width, and a six-axis external wrench expressed in the world frame~\cite{G02}. FTP-1 lets action queries attend to tactile and vision--language streams, with shared tactile processing and sensor-specific encoders trained on target demonstrations~\cite{G04}. These mechanisms make contact information available to action prediction.

Reactive Diffusion Policy (RDP) updates actions during a chunk: slow latent-action diffusion is paired with a fast GRU decoder receiving the latest tactile or force observation~\cite{U11}. This specialist imitation-learning method demonstrates feedback within ground contact tasks, complementing RTC's continuity between chunks. It does not impose a force bound.

Feedback control can complement these policies by shaping the robot's response to contact. Impedance control relates motion error to wrench, while admittance control converts wrench into motion adjustments~\cite{V01}. For floating robots, both must account for base response and changing contact constraints. ETS-VII used delayed-image compensation and speed limits for approach, followed by force-feedback compliance, capture, and braking~\cite{U12}. The target carried visual markers and moved within the clearance of a partially released docking mechanism; chaser attitude control was disabled. This orbital experiment illustrates the transition from visual approach to controlled contact.

\subsection{Constrained execution and runtime monitoring}

The execution layer checks command frames, units, limits, and observation age before acting. GR-2's ground deployment feeds optimized Cartesian trajectories to a 200\,Hz whole-body controller~\cite{G11}; this rate is distinct from model inference frequency. In an ISS experiment with Astrobee, a specialized neural network initializes trajectories and sequential convex programming handles the adopted dynamics and safety constraints~\cite{A05}.

Runtime assurance adds a mechanism for intervention. Abate et al. combine reachable-set overapproximations with a backup controller to enforce a line-of-sight constraint, validated on the ground ASTROS platform~\cite{U07}. The guarantee depends on the adopted dynamics, bounded disturbances, actuator limits, and verified backup set. Estimator covariance~\cite{V13} does not itself provide the required disturbance bound. Applying this approach to satellite capture would require corresponding contact and momentum models and validated recovery.

\begin{figure}[!htbp]
\centering
\includegraphics[width=\linewidth]{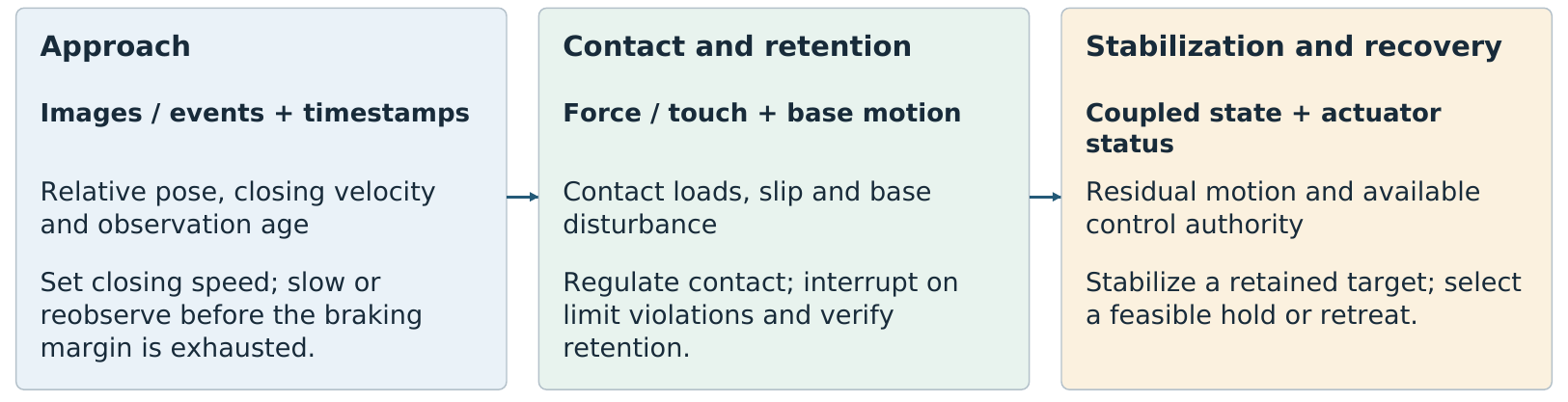}
\caption{A proposed capture example linking observations to execution and recovery. The phases draw on orbital approach/contact control and supervised execution~\cite{U12,U07,V13}; the combined foundation-model system is a design synthesis. Image semantics can guide target selection, while geometric and contact estimates determine admissible motion. A hold or retreat must remain feasible under the current dynamics and contact constraints.}
\label{fig:capture-feedback}
\end{figure}

For an approach, clearance must accommodate pose error, closing motion during sensing and computation, tracking error, and braking distance. Faster event acquisition helps only if encoding, inference and command execution preserve that timing benefit. As the remaining margin shrinks, the supervisor must reduce speed or obtain another observation while recovery remains feasible. After contact, the relevant conditions change to grip retention, contact load and combined motion. Figure~\ref{fig:capture-feedback} connects these decisions across one capture attempt; Section~\ref{sec:validation} turns them into staged measurements and transfer tests.

\section{Learning, adaptation, and onboard deployment}
\label{sec:adaptation}

\subsection{Choosing what to adapt}

Space adaptation addresses appearance, sensing and action consequences, with uneven data coverage: Open X-Embodiment contains diverse terrestrial trajectories but does not cover floating-base interaction~\cite{N25}. Space-LLaVA adapts visual--language knowledge and SPADES supplies spacecraft events; neither provides complete servicing trajectories~\cite{F02,N05}. DreamGen illustrates how generated videos broaden visual scenarios, while physical consistency and executable actions still require checking~\cite{G12}.

Octo changes observation and action interfaces; OpenVLA limits trainable parameters through low-rank adaptation~\cite{F15,F14}. New sensors or command conventions may require interface updates; changed inertia or contact may require dynamics or policy adaptation. Section~\ref{sec:validation} describes comparison controls. Ground training, pre-mission adaptation and onboard updates also have different computing and data budgets.

\subsection{Reducing deployment cost}

A VLM can supply spacecraft pseudo-labels for a smaller detection or segmentation model, leaving only the student at deployment~\cite{E03}. The cited study evaluates image datasets. Retaining the backbone preserves its inference and activation-memory costs, even with low-rank adaptation. Astrobee's specialist flight policy uses two 64-unit hidden layers within a sub-1~MB policy budget, separate from total avionics memory~\cite{V17}.

Physical Jetson AGX Orin profiling gives QwenVLA/OpenVLA-OFT peak memory of 7.39/19.20~GB, with separately evaluated LIBERO success of 78.8/95.3\%~\cite{T14}. Chunk latency is measured over 100 fixed-image/prompt queries under several power modes. BitVLA combines a pretrained ternary language backbone with quantization-aware visual distillation, reporting 1.4~GB model memory and 96.0\% LIBERO success~\cite{T15}. Its 73~ms per 25-action chunk is measured on A100; physical manipulation uses remote A800 inference. Thus compression, runtime memory and response time require separate comparisons on the intended hardware.

Inference optimization introduces a further tradeoff between output throughput and fresh feedback. In the OpenVLA-OFT ablation, parallel decoding raises A100 throughput from 4.2 to 15.9 generated actions/s; adding eight-action chunking raises it to 108.8 actions/s~\cite{U05}. Query latency nevertheless increases from 62.9 to 73.5~ms when chunking is added. These measurements average 100 queries and concern ablation variants, not the final continuous-action L1 configuration. Because each chunk executes fully before replanning, more generated actions do not imply faster responses to new observations. FAST likewise improves action encoding without measuring onboard energy savings~\cite{G08}.

\subsection{Spiking computation for local autonomy}

SNNs are relevant when a robot must sustain local sensing or control within power and heat limits during communication gaps. Stateful neurons and sparse communication offer a route to compact temporal processing~\cite{L01,L02}. One possible design couples a local spiking estimator or controller to a slower foundation model, using each at the rate its task requires.

Implementation can convert ANN activations into spikes or train spiking dynamics directly with surrogate gradients~\cite{L02}. Both routes require matching neuron operations, precision, connectivity and communication to the processor~\cite{L01}.

Quantized keypoint networks on Akida feed geometric PnP pose recovery using synthetic SPADES data and ground-truth crops~\cite{N07}. For event-to-frame (E2F) input, physical V1 reports 0.919\,W dynamic power at 16 FPS, versus 2.519\,W at 330 FPS for the floating-point Jetson Orin Nano baseline. Dividing power by throughput gives 57.4 versus 7.63\,mJ/frame: lower power accompanies higher dynamic inference energy at these operating points. On the full 35,889-input test set, quantization raises the E2F mean pose score (lower is better) from 0.0361 to 0.1009 and rejected poses from 1 to 162; those means exclude rejections. Hardware benchmarking uses 4,000 inputs per representation, distinct from the full-test accuracy comparison. V2 uses a heatmap architecture evaluated in cloud simulation, with higher-frequency throughput projected. Whole-system task accounting must also specify sensing, target detection, frame accumulation, PnP, host and idle costs.

For a wall-mounted cabin manipulator, a directly trained spiking actor--critic combines geometric states, tactile features and semantic cues; a PPO curriculum progresses from approach to grasp and lift~\cite{N21}. These Isaac Gym experiments test the complete policy, while energy remains an operation-count estimate.

Terrestrial VLAs extend this approach. SpikeVLA converts vision and language components and trains a population-coded action network with PPO and surrogate gradients~\cite{U01}. Spike-driven VLA combines trainable spiking vision, frozen spiking BERT, token fusion and state-conditioned action chunking~\cite{U02}. With fusion and decoding fixed, replacing its ANN encoders with spiking encoders lowers LIBERO success from 96.3\% to 92.4\% and estimated arithmetic energy from 179.7 to 20.9\,mJ per twelve-action chunk. The estimate includes the frozen encoder but excludes memory and other hardware overhead; encoder capacity and pretrained priors also change.

SpikingVLA converts NaVILA using dendritic channel mixing and adaptive firing~\cite{U03}. Internal steps propagate asynchronously but process the same observation; eight rather than four improve navigation while increasing estimated energy. All three use simulation and operation-based estimates, requiring hardware measurements.

\subsection{Continual learning and recovery}

MuSe uses replay during force-feedback adaptation, with storage and data-provenance costs~\cite{G14}. The space approach in Section~\ref{sec:planning} selectively updates components after simulated failures; its authors report that online optimization exceeds the power constraints of space-grade embedded computing~\cite{F07}.

A changed payload may require only a revised estimate, without backbone retraining. Ground updates allow regression testing, while onboard training competes with inference. After rollback, restored parameters, recurrent memory and state estimates must be reconciled with the robot's current physical state. Evaluation should measure retained and regained performance, update cost and switching behavior, beyond eventual policy reward.

\subsection{Thermal operation and energy measurement}

In vacuum, heat leaves onboard electronics through conduction and radiation~\cite{V19}. Thermal throttling or reduced duty cycles can slow inference and leave controllers acting on older observations. Deployment therefore requires sustaining the task's feedback rate within thermal limits.

Memory access, routing, neuron-state maintenance and input bursts also consume energy~\cite{L01}. NeuroBench separates algorithm complexity from system measurements, including preprocessing and postprocessing~\cite{U06}. Table~\ref{tab:energy-boundaries} distinguishes these measurement boundaries, from arithmetic estimates to proposed complete-task comparisons.

\begin{table}[!htbp]
\centering
\small
\setlength{\tabcolsep}{4pt}
\caption{Resource and energy evidence with different quantities and measurement scopes, not a cross-platform ranking. The final row proposes complete-task accounting. Akida comparison values use dynamic power divided by throughput and exclude idle energy.}
\label{tab:energy-boundaries}
\begin{tabularx}{\linewidth}{@{}YYYY@{}}
\toprule
Evidence type & Example & Quantities and units & Exclusions or conditions \\
\midrule
Runtime profiling & Jetson AGX Orin~\cite{T14} & Peak memory (GB); chunk latency; 15/30/50~W and MAX modes & Power modes specify limits, not measured draw \\
Arithmetic estimate & Manipulation SNN~\cite{N21} & Operations; assumed J/operation; estimated J/inference & No measured memory, routing, or idle cost \\
Power-derived inference energy & Akida V1 / Jetson Orin Nano~\cite{N07} & E2F: 57.4 / 7.63~mJ/frame, from dynamic power / FPS & AKD1000 chip / Jetson board; different precision and operating points \\
Measured board power & Nahuku, preprint Table 2~\cite{N08} & Idle: 0.935--0.938~W; running-minus-idle: 7--12~mW & Preloaded events; excludes streaming I/O and preprocessing \\
Proposed task accounting & Matched-task test & Integrated energy (J); J/completed task; completion rate & Declare sensors, host, recovery and actuators included \\
\bottomrule
\end{tabularx}
\end{table}

\begin{samepage}
The hardware configuration matters. Nahuku measurements use preloaded inputs, whereas the flight system used Kapoho Bay; the running-minus-idle increment is distinct from total power and the study's separately defined dynamic-power component~\cite{N08}. Task-level accounting should include failures and idle periods and specify actuator coverage.\par
\end{samepage}

\begin{samepage}
\subsection{Radiation tolerance and model robustness}

Radiation can corrupt computation even when the input is familiar. Avionics protections address different failures and impose different costs~\cite{V20}. Error-correcting codes cover bounded storage-error patterns; scrubbing consumes bandwidth and leaves intervals between checks; watchdogs detect stalled execution but may miss incorrect outputs. Replication adds power, synchronization and voting costs while remaining vulnerable to common failures. RedNet instead allocates protection according to layer sensitivity and spatially correlated upsets in satellite DNN workloads~\cite{P09}. Its payload and radiation-emulator results motivate selective protection as an alternative to uniform redundancy.\par
\end{samepage}

Vision-transformer studies identify vulnerable modules through soft-error experiments, including bit-flip injection~\cite{N17,N22}. Applying these results to control requires matching the fault model to the device and tracing output errors into actions. Compression or adaptation can change numerical sensitivity and fault propagation, requiring protection reassessment. Section~\ref{sec:validation} connects irradiation and closed-loop tests to the operational questions: is corruption detected before actuation, and does restart restore coherent control?


\section{Simulation, physical validation, and evaluation}
\label{sec:validation}

Validation must connect model behavior to the hardware, environment and task stage.

\subsection{From benchmark performance to operational claims}

Existing benchmarks cover different parts of the autonomy problem. CALVIN tests language-conditioned manipulation sequences and LIBERO studies transfer and continual learning~\cite{A10,A11}; neither represents free-floating reaction motion. SIMPLER examines whether simulation reflects physical policy performance, a relationship that needs reassessment when inertia or contact changes~\cite{A12}. Space perception resources provide complementary tests: SPEED+ addresses synthetic-to-laboratory visual transfer, SHIRT supports temporal pose tracking, and SpaceSense-Bench broadens synthetic multimodal perception~\cite{V12,V13,P08}. For task-level simulation, Space Robotics Bench supplies learning environments and Basilisk supplies modular astrodynamics~\cite{V15,V10}. Their usefulness depends on whether the implemented mechanisms reproduce those governing the chosen task.

Event simulation also needs sensor models. ESIM couples adaptive rendering with event generation~\cite{N23}; v2e models threshold mismatch, finite intensity-dependent bandwidth and noise~\cite{N24}. ELOPE combines PANGU imagery and v2e events with range and inertial measurements for lunar-descent estimation before touchdown~\cite{N18}. A closed-loop alternative regulates thrust using optical-flow divergence from tracked events~\cite{N19}. Success means reaching 50~m altitude; near-surface oscillations and unrealistic thrust requests remain. Both routes depend on rendering cadence, illumination, thresholds, noise and timestamps; event generation cannot recover information absent from source images.

\begin{samepage}
Sensor tests measure event statistics and motion-estimation error; task tests reveal their effects on approach or recovery. Timing must include accumulation, buffering and execution, which can consume asynchronous sensing's latency advantage.\par
\end{samepage}

Reaching a grasp pose, establishing contact, retaining a target and stabilizing the coupled system are distinct achievements. The dual-arm learning study stops short of post-contact force control~\cite{P02}, while Lunar-Bench tests reasoning about described scenarios~\cite{F08}.

The spacecraft label-distillation study splits SPEED+'s original test domains for adaptation and evaluation~\cite{E03}, changing the benchmark's untouched-domain protocol. Interpreting transfer therefore requires knowing which images, objects and trajectories were available during pretraining, pseudo-label generation, adaptation and model selection.

\subsection{From simulation through ground tests to space}

Ground testbeds preserve selected physical mechanisms while introducing their own constraints. SPICE-HL3 provides teleoperated rover recordings under LunaLab's lunar-analogue lighting, supporting optical validation separately from reduced-gravity dynamics~\cite{T02}. Planar air bearings permit relative motion and in-plane coupling; spherical arrangements permit attitude motion within balancing and travel limits~\cite{V01}. Support hardware affects the resulting dynamics. Correcting its added inertia improves manipulator vibration identification, while friction and air resistance still affect damping~\cite{P06}. A transferred policy may consequently learn to compensate for the apparatus as well as the intended robot dynamics.

Gravity offloading supports space mechanisms during ground tests by compensating for their weight. The cable-driven Motion Suspension System supports three-dimensional CAESAR arm motion and contact experiments~\cite{P04,P05}, but suspension forces, attachment geometry, tracking error and bandwidth still shape the motion. Robotic hardware-in-the-loop (HIL) testing combines modeled motion with physical sensing or interaction: EPOS supports proximity-operation testing and OOS-SIM includes a real manipulation arm~\cite{V06,V08}. Here contact stability depends on the rendered dynamics and delay; the cited EPOS stability experiment is one-dimensional~\cite{V07}. Table~\ref{tab:validation-media} identifies calibration quantities that help assess these differences for a chosen task.

\begin{table}[!htbp]
\centering
\small
\setlength{\tabcolsep}{3pt}
\caption{Validation media, preserved mechanisms and residual differences. Calibration quantities are proposed reporting items for the chosen task; environmental tests address separate conditions.}
\label{tab:validation-media}
\begin{tabularx}{\linewidth}{@{}>{\raggedright\arraybackslash}p{0.16\linewidth} Y Y Y >{\raggedright\arraybackslash}p{0.11\linewidth}@{}}
\toprule
Medium & Task or mechanism & Calibration quantities & Residual difference & Sources \\
\midrule
Digital simulation & Approach, sensing, contact models & Integrator step; sensor clocks; contact parameters & Model omissions and numerical error & \cite{V10,V15,N23} \\
Air bearings & Planar capture or attitude tests & Support inertia; drag; level/centre-of-mass offset & Constrained degrees of freedom & \cite{V01,P06} \\
Gravity offloading & Spatial arm motion and contact & Residual load; tracking error; bandwidth & Suspension constraints and forces & \cite{P04,P05} \\
Robotic HIL & Relative motion and physical contact & Delay; tracking; stiffness/damping & Rendered dynamics and workspace limits & \cite{V07,V08} \\
Processor HIL & Real inference with simulated robot & Execution latency; I/O; numerical format & Simulated embodiment and environment & \cite{V18} \\
Environmental tests & Thermal load or radiation response & Temperature/power; particle exposure/dose; recovery time & Configuration-specific, not task assurance & \cite{V19,P10,P12} \\
Flight experiments & ETS-VII orbital arm; ISS cabin free-flyer & State-estimation error; control timing; disturbances & Different environments and tested tasks & \cite{P07,A05} \\
\bottomrule
\end{tabularx}
\end{table}

\begin{figure}[!htbp]
\centering
\includegraphics[width=\linewidth]{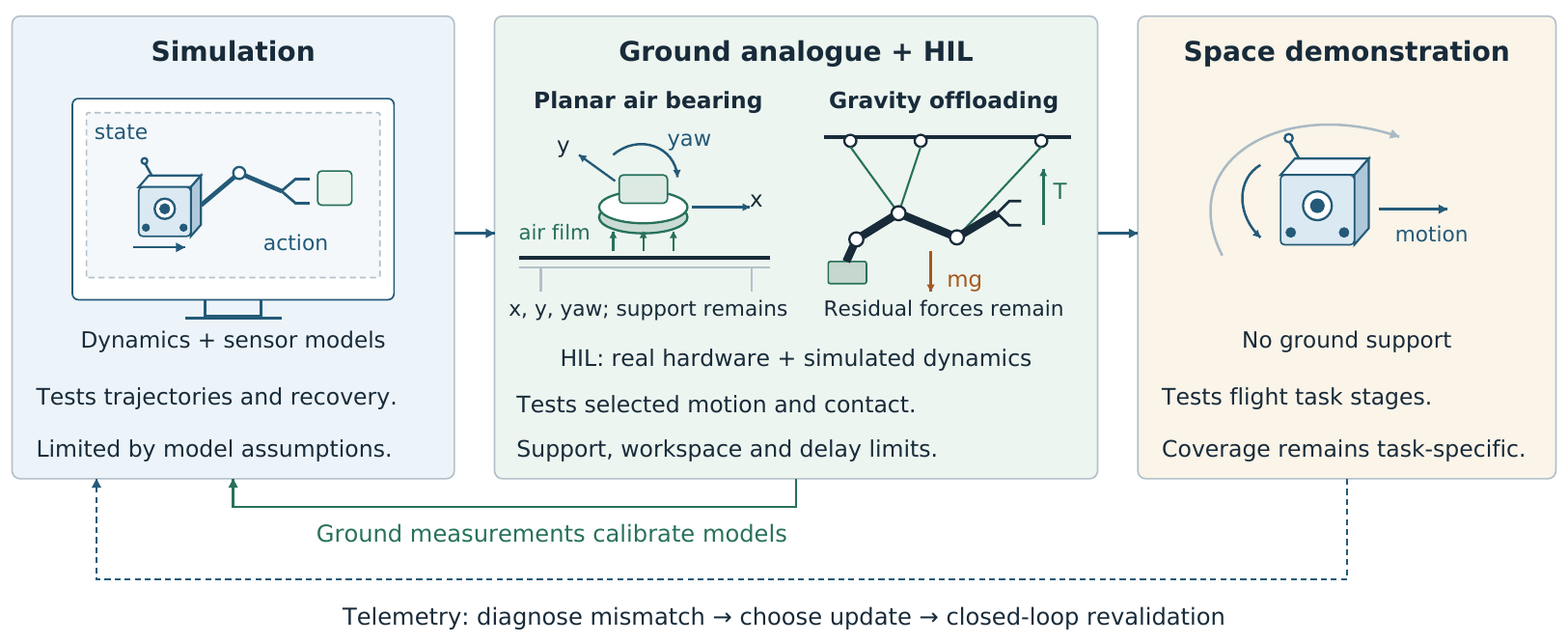}
\caption{Proposed sim-to-ground-to-space workflow and illustrative test mechanisms. Planar air bearings permit $x/y$ translation and yaw while retaining support; cable tension $T$ counteracts weight $mg$, with residual forces~\cite{V01,P04,P05}. Robotic HIL couples simulated motion to physical sensing or contact; processor HIL runs inference hardware with a simulated robot~\cite{V07,V08,V18}. Ground measurements calibrate models; telemetry informs diagnosis, update choice and closed-loop revalidation. Flight evidence remains task-specific.}
\label{fig:sim-ground-space}
\end{figure}

We propose an iterative sim-to-ground-to-space pathway (Figure~\ref{fig:sim-ground-space}). Ground analogues and HIL test selected dynamics, contact, sensor timing and computing costs against simulation. Compare generated events with physical recordings under matched motion and illumination, including buffering and noise; processor HIL connects SNN execution latency and energy to control behavior~\cite{V18}. Initially freeze the policy to expose transfer discrepancies before adaptation.

Diagnosis precedes the update choice: align commands, observations, actuator states and timestamps to separate dynamics mismatch from sensing, timing or facility effects. Distinguish simulator recalibration, ground retraining followed by upload, and onboard weight updates. Historical telemetry replay checks consistency with recorded trajectories; once a changed policy selects different actions, those trajectories cannot establish its closed-loop behavior. Revalidate the revised model or controller in calibrated closed-loop simulation and relevant HIL against both the observed discrepancy and the original operating envelope, using evaluation trajectories separate from calibration or adaptation.

Space demonstrations would progressively test representative task stages. Ground agreement during approach cannot establish contact or stabilization performance in space. Following NASA-STD-7009B, distinguish numerical verification, calibration and empirical validation; report residual uncertainty and extrapolation beyond tested inertia, contact and delay ranges~\cite{U13}.

LunaLab's fully randomized octree-based RL policy achieved 8/25 physical grasps versus 2/25 with reduced randomization~\cite{V16}. Each was selected from three training seeds, so these episodes do not measure variation across trained deployments. Full randomization reduced training stability; reduced randomization more often produced repeated poorly positioned grasps and timeouts. Simulation used terrestrial gravity (9.807~m/s$^2$): this tests visual and manipulation transfer, leaving lunar gravity and free-capture momentum management unresolved.

\subsection{Interpreting flight and hardware evidence}

Flight and processor tests answer different questions. ETS-VII provides orbital manipulation evidence~\cite{P07}, while Astrobee demonstrates task-specific learned control and trajectory warm starts aboard the ISS~\cite{V17,A05}. Falcon Neuro adds sensor flight evidence: its ISS-mounted nadir camera recorded Earth's surface in January 2022, allowing coastlines and clouds to be reconstructed from events~\cite{T04}.

LeLaR provides a specialist flight example of discrepancy diagnosis and supervised control. Its InnoCube controller issued three reaction-wheel torque commands at 1\,Hz using gyro-integrated relative attitude~\cite{C01}. Telemetry exposed startup delays up to approximately 20~s and intermittent unresponsiveness near zero speed; comparing commands, wheel speeds and attitude distinguished delayed actuation from measurement spikes. The authors retained the training wheel model to assess robustness against these effects. An independent monitor was configured to stop control and enter command-free idle on limit violations, followed by delayed wheel power-off and operator reset. No monitor intervention occurred in the reported experiments. The idle design relied on InnoCube remaining power-positive while tumbling; a robot in contact may instead need controlled holding or retreat.

The Loihi~2 control study converts a ReLU policy into a quantized sigma--delta network that transmits activation changes~\cite{V18}. A physical processor exchanges observations and actions with simulated Astrobee in Isaac Lab. It reduces measured inference energy relative to the GPU ANN, but increases final position and orientation errors. This connects deployment efficiency to control quality, while the related ANN's ISS flight remains a separate experiment. ISS cabin operation also differs from external vacuum and contact with uncontrolled satellites.

\begin{figure}[!htbp]
\centering
\includegraphics[width=\linewidth]{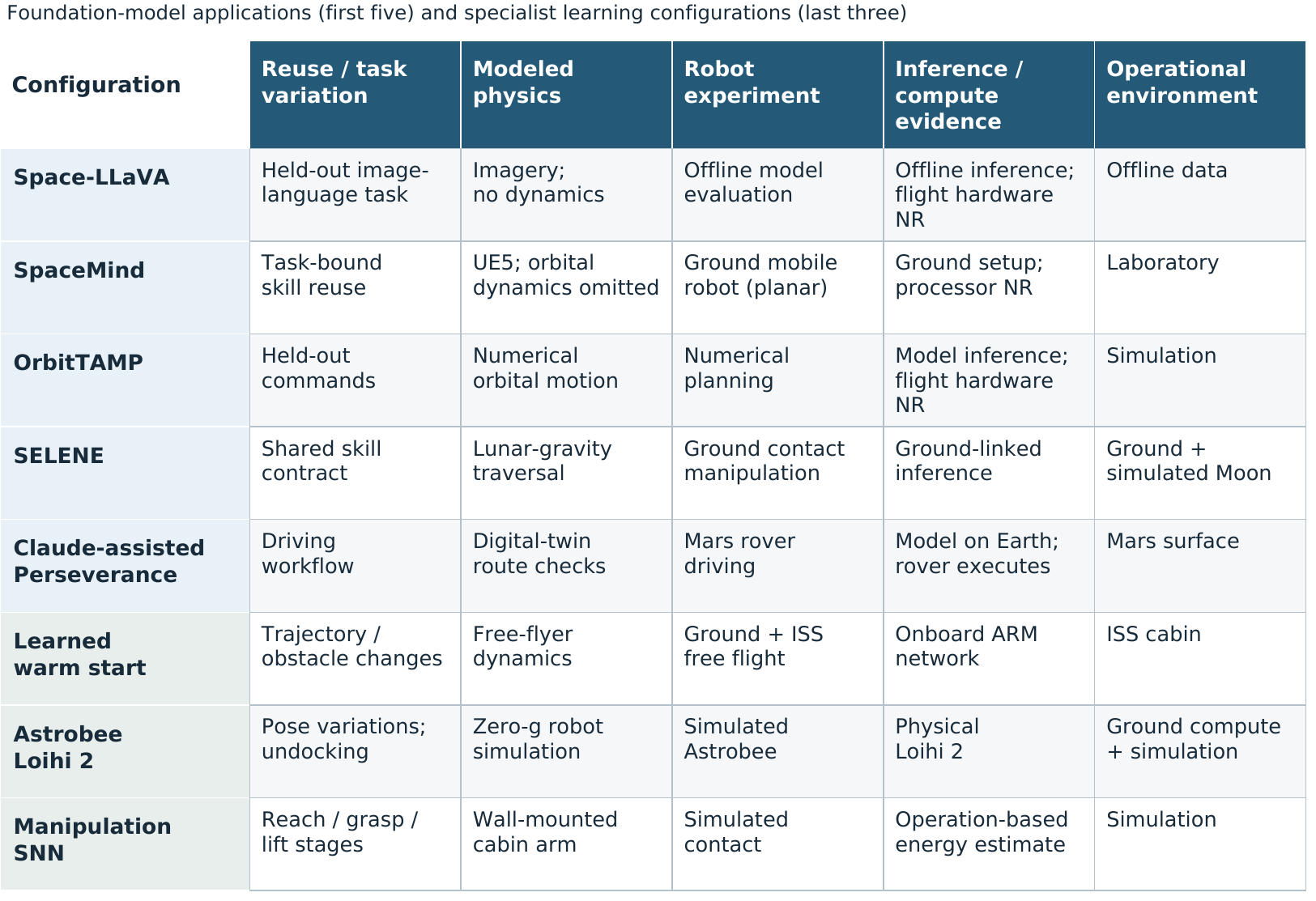}
\caption{Evidence for tested model--hardware configurations. Physical manipulation, free flight and rover driving are distinguished. Mars planning ran on Earth; Space-LLaVA's separate GPT-4o demonstration is excluded. NR denotes unreported information. Cell-level sources and qualifications accompany the manuscript~\cite{F02,F03,E05,F09,F12,A05,V18,N21}.}
\label{fig:evidence}
\end{figure}

\begin{samepage}
Figure~\ref{fig:evidence} separates language-task variation, skill reuse and physical execution across tested configurations. SELENE, for example, evaluates physical manipulation and simulated traversal through separate pathways. These capabilities do not form a single readiness score. No thermal-vacuum or irradiation tests of the configurations shown are reported.\par
\end{samepage}

\subsection{Environmental and resource testing}

Thermal-vacuum tests should run representative workloads with the intended conductive and radiative interfaces, recording temperature, power, sustained latency and task behavior to establish how heat affects feedback timing.

Radiation studies illustrate why the device and fault mechanism must also be specified. RedNet uses a hardware-in-the-loop radiation-error emulator~\cite{P09}; proton irradiation of Tensil reveals service interruption and silent incorrect outputs~\cite{P10}; ODIN combines neutron exposure with calibrated injection~\cite{P11}. Loihi has proton single-event-effect and total-dose measurements~\cite{P12}, but these cannot establish tolerance of the Loihi~2 configuration used in the control study~\cite{V18}. To connect such computing tests to an embodied system, experiments must trace fault detection, rejected commands, recovery duration and the physical consequences of interruption.

Use irradiation data to calibrate device-specific fault models, then inject representative faults during execution. Report exposure, numerical configuration, monitored outputs and interventions: unchanged success can conceal repeated supervisory recovery, while rare errors may cause consequential actions.

\subsection{A reusable evaluation protocol}

We propose a satellite-capture experiment crossing seen/unseen task compositions with seen/unseen physical conditions. Group tasks by target, grasp site and approach--retreat--recapture sequence, and physics by inertia, target rotation and contact parameters. Evaluate four cells: both groups seen, unseen tasks only, unseen physics only, and both unseen. These labels describe the initial documented development split; report subsequent adaptation exposure and known pretraining overlap or provenance uncertainty separately. Calibrate physical ranges and thresholds against the intended mission.

Use two evaluation modes. \emph{Held-out generalization} fixes the checkpoint across test conditions: report the original model and a version adapted only on the seen/seen set, reserving unseen groups for evaluation. \emph{Target-physics adaptation} allows a separate adaptation set from new physical conditions and measures performance against interaction and compute budgets. Split adaptation, model-selection and final evaluation trajectories before augmentation; final evaluation data must influence neither adaptation nor model selection. Log model/simulator versions, changed components, inference location and numerical format.

Within a route, isolate model changes with sensors, action interfaces, executor and supervision fixed; test controller changes separately. Compare adaptation methods from the same pretrained checkpoint; isolate pretraining with a same-architecture random-initialization control and matched downstream data and optimization budgets where feasible. Across routes, compare foundation-model, specialist and conventional systems on the same robot, tasks, available observations, physical conditions, intervention criteria and resource budgets, using paired initial conditions. Retain each route's necessary interfaces and account for its controller, latency and recovery costs.

Score approach, contact, retention and stabilization separately, with task-calibrated limits on relative-state error, contact peaks, retention duration and residual motion. Report uninterrupted completion separately from recovery, including fault onset, deadlines, rejected actions, retreats, operator corrections and resets. Measure base disturbance, constraint violations, latency distributions and energy over declared components, including idle periods and failed attempts. Separate independent training/adaptation runs from evaluation episodes; retain task/scenario groups and use interval estimates suited to the sampling design~\cite{U14}. Successive actions are not independent trials. Retain unrecovered trials.


\section{Synthesis and research priorities}
\label{sec:discussion}

Table~\ref{tab:representative-evidence} separates reused components, target-domain updates, selected comparison conditions, and the interventions whose benefits were measured.

\begin{table}[!htbp]
\centering
\small
\setlength{\tabcolsep}{3pt}
\caption{Within-study comparisons of reuse and measured contributions. Conditions refer to selected experiments. NR: not reported.}
\label{tab:representative-evidence}
\begin{tabularx}{\linewidth}{@{}>{\raggedright\arraybackslash}p{0.14\linewidth} Y Y Y@{}}
\toprule
Study & Reuse and target update & Comparison conditions & Finding and attribution \\
\midrule
Cosmos Policy~\cite{G13} & Video-model initialization; robot policy training & LIBERO; pretrained versus random initialization; same gradient steps & 98.5 versus 94.6\% success. Video-pretraining benefit at the tested training budget \\
SpaceMind~\cite{F03} & Frozen VLM; skill instructions updated over five rounds per group & Six task-bound groups; no cross-task skill transfer & Four groups improve; instruction-based recovery within repeated tasks \\
OrbitTAMP~\cite{E05} & Frozen LLM; prompt developed on 100 commands & Held-out instance/radio-style sets, 100 commands each; same ontology & Exact partial MissionConfig: rules 92/57\%; Qwen3.5-9B 78/70\%; Terra 98/98\%. Parser contribution \\
Spike-driven VLA~\cite{U02} & Pretrained vision/language encoders; language frozen; 80k policy updates & LIBERO tasks used in training; demonstration count NR & ANN/SNN: 96.3/92.4\% success; estimated 179.7/20.9 mJ/chunk. Priors and capacity differ \\
OpenVLA-OFT~\cite{U05} & OpenVLA checkpoint; LoRA on filtered trajectories from 500-demo suite pools & LIBERO execution after adaptation; fixed-query A100 timing & Adding chunking: 15.9 $\rightarrow$ 108.8 actions/s, 62.9 $\rightarrow$ 73.5 ms/query. Decoding benefit \\
\bottomrule
\end{tabularx}
\end{table}

\subsection{What the current results support}

Terrestrial comparisons support specific pretraining benefits: video initialization improves Cosmos Policy at matched gradient steps~\cite{G13}, while OpenVLA's robot-trajectory pretraining improves over directly adapting Prismatic, with vision--language pretraining retained in both~\cite{F14}. These benefits need testing under space dynamics and scarce interaction data.

Other studies isolate interface or adaptation effects: OrbitTAMP compares instruction parsers with planning fixed~\cite{E05}, SpaceMind revises task-specific recovery instructions~\cite{F03}, and dynamics updates address changed action consequences~\cite{F07}. OpenVLA-OFT increases throughput and query latency; spiking-encoder comparisons change capacity and priors alongside arithmetic cost~\cite{U05,U02}. These interventions require different explanations of the measured gain.

In the capture example (Figure~\ref{fig:capture-feedback}), temporal sensing must deliver usable state in time, and computing savings must survive task execution. Astrobee and LeLaR provide specialist flight-control experience; Loihi~2 couples a physical processor to a simulated robot~\cite{V17,A05,C01,V18}. Joint tests of task reuse, coupled dynamics, sustained computation and recovery remain the main gap for foundation-model systems.

\subsection{Research questions that follow from these comparisons}

\paragraph{Where should physical adaptation occur?}
Use the crossed capture tests (Section~\ref{sec:validation}) to separate held-out generalization from target-physics adaptation. Compare state conditioning, policy updates and controller changes at matched budgets as inertia or contact varies.

\paragraph{Does transfer survive contact and recovery?}
Test held-out task--physics combinations through contact, retention and stabilization, retaining failed attempts to identify whether recovery depends on the policy, controller or operator.

\paragraph{Do efficiency gains survive deployment?}
Compare sensing and computing variants on matched tasks, including preprocessing, memory, idle periods and recovery. Diagnose transfer discrepancies before choosing an update and test it in the closed loop. Report feedback timing, task energy and sustained thermal performance alongside accuracy.

\subsection{Conclusion}

Foundation models contribute reusable representations, task interfaces and learned behaviors to autonomous robotics, with controlled ground studies supporting specific pretraining benefits. The central space challenge is preserving those benefits through changed dynamics, contact and limited onboard resources. Event sensing and spiking computation offer complementary design options; sim-to-ground-to-space validation provides a path to test them. Separating generalization from bounded target-domain adaptation would connect model reuse to complete-task performance, recovery and resource needs.


\FloatBarrier

\begingroup
\fontsize{9}{9.6}\selectfont
\raggedright
\setlength{\bibsep}{0pt plus 0.5pt}

\endgroup

\begin{thebibliography}{120}
\providecommand{\natexlab}[1]{#1}
\providecommand{\url}[1]{\texttt{#1}}
\expandafter\ifx\csname urlstyle\endcsname\relax
  \providecommand{\doi}[1]{doi: #1}\else
  \providecommand{\doi}{doi: \begingroup \urlstyle{rm}\Url}\fi

\bibitem[Papadopoulos et~al.(2021)Papadopoulos, Aghili, Ma, and
  Lampariello]{V01}
Evangelos Papadopoulos, Farhad Aghili, Ou~Ma, and Roberto Lampariello.
\newblock {Robotic Manipulation and Capture in Space: A Survey}.
\newblock \emph{Frontiers in Robotics and AI}, 8:\penalty0 686723, 2021.
\newblock \doi{10.3389/frobt.2021.686723}.

\bibitem[Nesnas et~al.(2021)Nesnas, Fesq, and Volpe]{V02}
Issa A.~D. Nesnas, Lorraine~M. Fesq, and Richard~A. Volpe.
\newblock {Autonomy for Space Robots: Past, Present, and Future}.
\newblock \emph{Current Robotics Reports}, 2:\penalty0 251--263, 2021.
\newblock \doi{10.1007/s43154-021-00057-2}.

\bibitem[Firoozi et~al.(2023)Firoozi, Tucker, Tian, et~al.]{A01}
Roya Firoozi, Johnathan Tucker, Stephen Tian, et~al.
\newblock {Foundation Models in Robotics: Applications, Challenges, and the
  Future}.
\newblock \href{https://arxiv.org/abs/2312.07843v1}{arXiv:2312.07843v1}, 2023.

\bibitem[Xu et~al.(2024)Xu, Wu, Wen, Li, Liu, Che, and Tang]{A03}
Zhiyuan Xu, Kun Wu, Junjie Wen, Jinming Li, Ning Liu, Zhengping Che, and Jian
  Tang.
\newblock {A Survey on Robotics with Foundation Models: toward Embodied AI}.
\newblock \href{https://arxiv.org/abs/2402.02385v1}{arXiv:2402.02385v1}, 2024.

\bibitem[Hu et~al.(2023)Hu, Xie, Jain, et~al.]{A04}
Yafei Hu, Quanting Xie, Vidhi Jain, et~al.
\newblock {Toward General-Purpose Robots via Foundation Models: A Survey and
  Meta-Analysis}.
\newblock \href{https://arxiv.org/abs/2312.08782v3}{arXiv:2312.08782v3}, 2023.
\newblock Revised 2024-10-01.

\bibitem[Huang et~al.(2026{\natexlab{a}})Huang, Chen, Hao, et~al.]{F01}
Zeyuan Huang, Gang Chen, Zixuan Hao, et~al.
\newblock {Artificial Intelligence-Enabled Space Robot Operations:
  Technologies, Challenges and Prospects}.
\newblock \href{https://arxiv.org/abs/2609.16880v1}{arXiv:2609.16880v1},
  2026{\natexlab{a}}.

\bibitem[Luo et~al.(2026)Luo, Zhang, and Wang]{E13}
Tao Luo, Yahang Zhang, and Yaobing Wang.
\newblock {Research Status and Application Prospects of Multimodal Large Models
  for Space Robotics}.
\newblock \emph{Spacecraft Engineering}, 35\penalty0 (1):\penalty0 128--136,
  2026.
\newblock \doi{10.3969/j.issn.1673-8748.2026.01.017}.
\newblock In Chinese; title translated.

\bibitem[Gallego et~al.(2022)Gallego, Delbruck, Orchard, et~al.]{N01}
Guillermo Gallego, Tobi Delbruck, Garrick Orchard, et~al.
\newblock {Event-Based Vision: A Survey}.
\newblock \emph{IEEE Transactions on Pattern Analysis and Machine
  Intelligence}, 44\penalty0 (1):\penalty0 154--180, 2022.
\newblock \doi{10.1109/TPAMI.2020.3008413}.

\bibitem[Li et~al.(2026)Li, Li, Glover, Fan, Li, Bartolozzi, Benosman, and
  Tian]{L01}
Jianing Li, Dianze Li, Arren Glover, Xiaopeng Fan, Guoqi Li, Chiara Bartolozzi,
  Ryad~B. Benosman, and Yonghong Tian.
\newblock {Neuromorphic Object Detection: An In-Depth Study and Future
  Directions}.
\newblock \emph{Proceedings of the IEEE}, 114\penalty0 (5):\penalty0 712--739,
  2026.
\newblock \doi{10.1109/JPROC.2026.3717677}.

\bibitem[Ma and Tian(2023)]{L02}
Zhengyu Ma and Yonghong Tian.
\newblock {Theories and Methods for Large-Scale Brain-Inspired Neural
  Networks}.
\newblock \emph{Chinese Science Bulletin}, 68\penalty0 (35):\penalty0
  4764--4781, 2023.
\newblock \doi{10.1360/TB-2023-0775}.
\newblock In Chinese; English title supplied by the authors.

\bibitem[Kawaharazuka et~al.(2024)Kawaharazuka, Matsushima, Gambardella, Guo,
  Paxton, and Zeng]{A02}
Kento Kawaharazuka, Tatsuya Matsushima, Andrew Gambardella, Jiaxian Guo, Chris
  Paxton, and Andy Zeng.
\newblock {Real-World Robot Applications of Foundation Models: A Review}.
\newblock \href{https://arxiv.org/abs/2402.05741v1}{arXiv:2402.05741v1}, 2024.

\bibitem[Bartolozzi et~al.(2022)Bartolozzi, Indiveri, and Donati]{N02}
Chiara Bartolozzi, Giacomo Indiveri, and Elisa Donati.
\newblock {Embodied neuromorphic intelligence}.
\newblock \emph{Nature Communications}, 13:\penalty0 1024, 2022.
\newblock \doi{10.1038/s41467-022-28487-2}.

\bibitem[Li and Tian(2021)]{L03}
Jianing Li and Yonghong Tian.
\newblock {Recent Advances in Neuromorphic Vision Sensors: A Survey}.
\newblock \emph{Chinese Journal of Computers}, 44\penalty0 (6):\penalty0
  1258--1286, 2021.
\newblock \doi{10.11897/SP.J.1016.2021.01258}.
\newblock In Chinese; English title supplied by the authors.

\bibitem[{NASA Small Spacecraft Systems Virtual
  Institute}(2026{\natexlab{a}})]{V19}
{NASA Small Spacecraft Systems Virtual Institute}.
\newblock {7.0 Thermal Control}.
\newblock State-of-the-Art of Small Spacecraft Technology, 2026 web edition,
  2026{\natexlab{a}}.
\newblock URL
  \url{https://www.nasa.gov/smallsat-institute/sst-soa/thermal-control/}.
\newblock Accessed 2026-10-15.

\bibitem[{NASA Small Spacecraft Systems Virtual
  Institute}(2026{\natexlab{b}})]{V20}
{NASA Small Spacecraft Systems Virtual Institute}.
\newblock {8.0 Small Spacecraft Avionics}.
\newblock State-of-the-Art of Small Spacecraft Technology, 2026 web edition,
  2026{\natexlab{b}}.
\newblock URL
  \url{https://www.nasa.gov/smallsat-institute/sst-soa/small-spacecraft-avionics/}.
\newblock Accessed 2026-10-15.

\bibitem[Yoshida(2003)]{P07}
Kazuya Yoshida.
\newblock {Engineering Test Satellite VII Flight Experiments for Space Robot
  Dynamics and Control: Theories on Laboratory Test Beds Ten Years Ago, Now in
  Orbit}.
\newblock \emph{The International Journal of Robotics Research}, 22\penalty0
  (5):\penalty0 321--335, 2003.
\newblock \doi{10.1177/0278364903022005003}.

\bibitem[Takubo et~al.(2026)Takubo, Gammelli, Pavone, and D'Amico]{E05}
Yuji Takubo, Daniele Gammelli, Marco Pavone, and Simone D'Amico.
\newblock {OrbitTAMP: Grounding Language Models for Task and Motion Planning in
  Spacecraft Rendezvous}.
\newblock \href{https://arxiv.org/abs/2610.01093v1}{arXiv:2610.01093v1}, 2026.

\bibitem[Park and D'Amico(2023)]{V13}
Tae~Ha Park and Simone D'Amico.
\newblock {Adaptive Neural Network-based Unscented Kalman Filter for Robust
  Pose Tracking of Noncooperative Spacecraft}.
\newblock \emph{Journal of Guidance, Control, and Dynamics}, 46\penalty0
  (9):\penalty0 1671--1688, 2023.
\newblock \doi{10.2514/1.G007387}.

\bibitem[Zhao et~al.(2024)Zhao, Wang, Fan, Gao, and Zhang]{P01}
Wenbo Zhao, Shengjie Wang, Yixuan Fan, Yang Gao, and Tao Zhang.
\newblock {SpaceOctopus: An Octopus-inspired Motion Planning Framework for
  Multi-arm Space Robot}.
\newblock \href{https://arxiv.org/abs/2403.08219v1}{arXiv:2403.08219v1}, 2024.

\bibitem[Cao et~al.(2022)Cao, Wang, Zheng, Ma, Xie, and Liu]{P02}
Yuxue Cao, Shengjie Wang, Xiang Zheng, Wenke Ma, Xinru Xie, and Lei Liu.
\newblock {Reinforcement Learning with Prior Policy Guidance for Motion
  Planning of Dual-Arm Free-Floating Space Robot}.
\newblock \href{https://arxiv.org/abs/2209.01434v1}{arXiv:2209.01434v1}, 2022.

\bibitem[Niksirat et~al.(2020)Niksirat, Daca, and Skonieczny]{U08}
Parna Niksirat, Adriana Daca, and Krzysztof Skonieczny.
\newblock {The effects of reduced-gravity on planetary rover mobility}.
\newblock \emph{The International Journal of Robotics Research}, 39\penalty0
  (7):\penalty0 797--811, 2020.
\newblock \doi{10.1177/0278364920913945}.

\bibitem[Park et~al.(2022)Park, M{\"{a}}rtens, Lecuyer, Izzo, and D'Amico]{V12}
Tae~Ha Park, Marcus M{\"{a}}rtens, G{\'{e}}rard Lecuyer, Dario Izzo, and Simone
  D'Amico.
\newblock {SPEED+: Next-Generation Dataset for Spacecraft Pose Estimation
  across Domain Gap}.
\newblock In \emph{IEEE Aerospace Conference}, 2022.
\newblock \doi{10.1109/AERO53065.2022.9843439}.

\bibitem[Burmeister et~al.(2018)Burmeister, Elgner, Preusker, Stark, and
  Oberst]{C03}
S.~Burmeister, S.~Elgner, F.~Preusker, A.~Stark, and J.~Oberst.
\newblock {Thermal Effects on Camera Focal Length in MESSENGER Star Calibration
  and Orbital Imaging}.
\newblock \emph{The International Archives of the Photogrammetry, Remote
  Sensing and Spatial Information Sciences}, XLII-3:\penalty0 103--105, 2018.
\newblock \doi{10.5194/isprs-archives-XLII-3-103-2018}.

\bibitem[Antonsanti et~al.(2022)Antonsanti, Virmontois, Lauenstein, Roch,
  Dewitte, and Goiffon]{C02}
Aubin Antonsanti, C{\'e}dric Virmontois, Jean-Marie Lauenstein, Alexandre~Le
  Roch, Hugo Dewitte, and Vincent Goiffon.
\newblock {Probing Dark Current Random Telegraph Signal in a Small Pitch
  Vertically Pinned Photodiode CMOS Image Sensor After Proton Irradiation}.
\newblock \emph{IEEE Transactions on Nuclear Science}, 69\penalty0
  (7):\penalty0 1506--1514, 2022.
\newblock \doi{10.1109/TNS.2022.3160056}.

\bibitem[Yu et~al.(2025)Yu, Liu, Yu, et~al.]{G02}
Jiawen Yu, Hairuo Liu, Qiaojun Yu, et~al.
\newblock {ForceVLA: Enhancing VLA Models with a Force-aware MoE for
  Contact-rich Manipulation}.
\newblock \href{https://arxiv.org/abs/2505.22159v3}{arXiv:2505.22159v3}, 2025.

\bibitem[Feng et~al.(2025)Feng, Hu, Xia, Gao, Shen, Sun, Fang, and Hu]{G03}
Ruoxuan Feng, Jiangyu Hu, Wenke Xia, Tianci Gao, Ao~Shen, Yuhao Sun, Bin Fang,
  and Di~Hu.
\newblock {AnyTouch: Learning Unified Static-Dynamic Representation across
  Multiple Visuo-tactile Sensors}.
\newblock \href{https://arxiv.org/abs/2502.12191v3}{arXiv:2502.12191v3}, 2025.

\bibitem[Wang et~al.(2024{\natexlab{a}})Wang, Chen, Zhao, and He]{G05}
Lirui Wang, Xinlei Chen, Jialiang Zhao, and Kaiming He.
\newblock {Scaling Proprioceptive-Visual Learning with Heterogeneous
  Pre-trained Transformers}.
\newblock \href{https://arxiv.org/abs/2409.20537v1}{arXiv:2409.20537v1},
  2024{\natexlab{a}}.

\bibitem[Memon et~al.(2026)Memon, Graczyk, Swako{\'{n}}, Grzanka, Kusyk, and
  Papadakis]{P10}
Saad Memon, Rafal Graczyk, Jan Swako{\'{n}}, Leszek Grzanka, Sebastian Kusyk,
  and Mike Papadakis.
\newblock {Proton Irradiation Characterization of an Open-Source ML Accelerator
  on a Zynq UltraScale+ MPSoC}.
\newblock \href{https://arxiv.org/abs/2609.05249v1}{arXiv:2609.05249v1}, 2026.

\bibitem[Nijsink et~al.(2026)Nijsink, Forlin, Yousefzadeh, and Ottavi]{P11}
Wim Nijsink, Bruno~Endres Forlin, Amirreza Yousefzadeh, and Marco Ottavi.
\newblock {Shooting Neutrons at Neurons: Radiation Testing of a Spiking Neural
  Network on Flash-Based FPGAs}.
\newblock \href{https://arxiv.org/abs/2605.00030v1}{arXiv:2605.00030v1}, 2026.

\bibitem[Schmaus et~al.(2019)Schmaus, Leidner, Bayer, Pleintinger, Kr{\"u}ger,
  and Lii]{U10}
Peter Schmaus, Daniel Leidner, Ralph Bayer, Benedikt Pleintinger, Thomas
  Kr{\"u}ger, and Neal~Y. Lii.
\newblock {Continued Advances in Supervised Autonomy User Interface Design for
  METERON SUPVIS Justin}.
\newblock In \emph{2019 IEEE Aerospace Conference}, 2019.
\newblock \doi{10.1109/AERO.2019.8741885}.

\bibitem[{Open X-Embodiment Collaboration}(2023)]{N25}
{Open X-Embodiment Collaboration}.
\newblock {Open X-Embodiment: Robotic Learning Datasets and RT-X Models}.
\newblock \href{https://arxiv.org/abs/2310.08864v9}{arXiv:2310.08864v9}, 2023.
\newblock Revised 2025-05-14.

\bibitem[Huang et~al.(2022)Huang, Xia, Xiao, et~al.]{A09}
Wenlong Huang, Fei Xia, Ted Xiao, et~al.
\newblock {Inner Monologue: Embodied Reasoning through Planning with Language
  Models}.
\newblock \href{https://arxiv.org/abs/2207.05608v1}{arXiv:2207.05608v1}, 2022.

\bibitem[Kim et~al.(2024)Kim, Pertsch, Karamcheti, et~al.]{F14}
Moo~Jin Kim, Karl Pertsch, Siddharth Karamcheti, et~al.
\newblock {OpenVLA: An Open-Source Vision-Language-Action Model}.
\newblock \href{https://arxiv.org/abs/2406.09246v3}{arXiv:2406.09246v3}, 2024.

\bibitem[Ahn et~al.(2022)Ahn, Brohan, Brown, et~al.]{F18}
Michael Ahn, Anthony Brohan, Noah Brown, et~al.
\newblock {Do As I Can, Not As I Say: Grounding Language in Robotic
  Affordances}.
\newblock \href{https://arxiv.org/abs/2204.01691v2}{arXiv:2204.01691v2}, 2022.

\bibitem[Driess et~al.(2023)Driess, Xia, Sajjadi, et~al.]{F19}
Danny Driess, Fei Xia, Mehdi S.~M. Sajjadi, et~al.
\newblock {PaLM-E: An Embodied Multimodal Language Model}.
\newblock \href{https://arxiv.org/abs/2303.03378v1}{arXiv:2303.03378v1}, 2023.

\bibitem[Banerjee et~al.(2025)Banerjee, Cauligi, and Pavone]{A05}
Somrita Banerjee, Abhishek Cauligi, and Marco Pavone.
\newblock {Deep Learning Warm Starts for Trajectory Optimization on the
  International Space Station}.
\newblock \href{https://arxiv.org/abs/2505.05588v3}{arXiv:2505.05588v3}, 2025.

\bibitem[Liang et~al.(2022)Liang, Huang, Xia, Xu, Hausman, Ichter, Florence,
  and Zeng]{A08}
Jacky Liang, Wenlong Huang, Fei Xia, Peng Xu, Karol Hausman, Brian Ichter, Pete
  Florence, and Andy Zeng.
\newblock {Code as Policies: Language Model Programs for Embodied Control}.
\newblock \href{https://arxiv.org/abs/2209.07753v4}{arXiv:2209.07753v4}, 2022.
\newblock Revised 2023-05-25.

\bibitem[Yang et~al.(2026)Yang, Tang, Ma, Fan, Wang, Li, and Yang]{T09}
Yaodong Yang, Hongyao Tang, Yi~Ma, Xingyu Fan, Weixun Wang, Jinpeng Li, and
  Tianpei Yang.
\newblock {Code Owns the Simulation, Jev Owns the Evaluation}.
\newblock \href{https://arxiv.org/abs/2610.01834v2}{arXiv:2610.01834v2}, 2026.
\newblock Revised 2026-10-10.

\bibitem[{Octo Model Team} et~al.(2024){Octo Model Team}, Ghosh, Walke,
  et~al.]{F15}
{Octo Model Team}, Dibya Ghosh, Homer Walke, et~al.
\newblock {Octo: An Open-Source Generalist Robot Policy}.
\newblock \href{https://arxiv.org/abs/2405.12213v2}{arXiv:2405.12213v2}, 2024.

\bibitem[Black et~al.(2024)Black, Brown, Driess, et~al.]{F16}
Kevin Black, Noah Brown, Danny Driess, et~al.
\newblock {$\pi_0$: A Vision-Language-Action Flow Model for General Robot
  Control}.
\newblock \href{https://arxiv.org/abs/2410.24164v4}{arXiv:2410.24164v4}, 2024.
\newblock Revised 2026-01-08.

\bibitem[Yang et~al.(2023)Yang, Du, Ghasemipour, Tompson, Kaelbling,
  Schuurmans, and Abbeel]{G10}
Sherry Yang, Yilun Du, Kamyar Ghasemipour, Jonathan Tompson, Leslie Kaelbling,
  Dale Schuurmans, and Pieter Abbeel.
\newblock {Learning Interactive Real-World Simulators}.
\newblock \href{https://arxiv.org/abs/2310.06114v3}{arXiv:2310.06114v3}, 2023.
\newblock Revised 2024-09-26.

\bibitem[Kim et~al.(2026)Kim, Gao, Lin, et~al.]{G13}
Moo~Jin Kim, Yihuai Gao, Tsung-Yi Lin, et~al.
\newblock {Cosmos Policy: Fine-Tuning Video Models for Visuomotor Control and
  Planning}.
\newblock \href{https://arxiv.org/abs/2601.16163v1}{arXiv:2601.16163v1}, 2026.

\bibitem[Abate et~al.(2024)Abate, Mote, Dor, et~al.]{U07}
Matthew Abate, Mark Mote, Mehregan Dor, et~al.
\newblock {Run Time Assurance for Spacecraft Attitude Control Under
  Nondeterministic Assumptions}.
\newblock \emph{IEEE Transactions on Control Systems Technology}, 2024.
\newblock \doi{10.1109/TCST.2023.3340624}.

\bibitem[Knestel et~al.(2025)Knestel, Mayershofer, K{\"{o}}pken, et~al.]{E01}
Philipp~G. Knestel, Luisa Mayershofer, Anne K{\"{o}}pken, et~al.
\newblock {NealAI: A RAG-based Chat Assistant Supporting Command of a Robotic
  Team from the ISS}.
\newblock In \emph{International Conference on Space Robotics (iSpaRo)}, pages
  684--691, 2025.
\newblock \doi{10.1109/iSpaRo66239.2025.11436289}.

\bibitem[Foutter et~al.(2024)Foutter, Gammelli, Kruger, Foss, Bhoj, Guffanti,
  D'Amico, and Pavone]{F02}
Matthew Foutter, Daniele Gammelli, Justin Kruger, Ethan Foss, Praneet Bhoj,
  Tommaso Guffanti, Simone D'Amico, and Marco Pavone.
\newblock {Space-LLaVA: a Vision-Language Model Adapted to Extraterrestrial
  Applications}.
\newblock \href{https://arxiv.org/abs/2408.05924v2}{arXiv:2408.05924v2}, 2024.
\newblock Revised 2025-01-18.

\bibitem[P{\'{e}}rez-Ramos et~al.(2026)P{\'{e}}rez-Ramos, Altamirano-Robles,
  and Ch{\'{a}}vez-Dagostino]{F10}
Cristina P{\'{e}}rez-Ramos, Leopoldo Altamirano-Robles, and Miguel
  Ch{\'{a}}vez-Dagostino.
\newblock {Zero-shot semantic landmark-based visual odometry using foundation
  models for unstructured planetary exploration}.
\newblock \emph{Frontiers in Robotics and AI}, 13:\penalty0 1819714, 2026.
\newblock \doi{10.3389/frobt.2026.1819714}.

\bibitem[Wen et~al.(2024)Wen, Yang, Kautz, and Birchfield]{U09}
Bowen Wen, Wei Yang, Jan Kautz, and Stan Birchfield.
\newblock {FoundationPose: Unified 6D Pose Estimation and Tracking of Novel
  Objects}.
\newblock In \emph{Proceedings of the IEEE/CVF Conference on Computer Vision
  and Pattern Recognition (CVPR)}, pages 17868--17879, 2024.
\newblock URL \url{https://arxiv.org/abs/2312.08344v2}.
\newblock Reviewed arXiv:2312.08344v2, revised 2024-03-26.

\bibitem[Rathinam et~al.(2024)Rathinam, Qadadri, and Aouada]{N05}
Arunkumar Rathinam, Haytam Qadadri, and Djamila Aouada.
\newblock {SPADES: A Realistic Spacecraft Pose Estimation Dataset using Event
  Sensing}.
\newblock In \emph{IEEE International Conference on Robotics and Automation
  (ICRA)}, 2024.
\newblock \doi{10.1109/ICRA57147.2024.10611231}.

\bibitem[Rathinam et~al.(2026)Rathinam, Lecomte, Reelsen, Lenz, von Arnim, and
  Aouada]{N07}
Arunkumar Rathinam, Jules Lecomte, Jost Reelsen, Gregor Lenz, Axel von Arnim,
  and Djamila Aouada.
\newblock {Efficient Onboard Spacecraft Pose Estimation with Event Cameras and
  Neuromorphic Hardware}.
\newblock \href{https://arxiv.org/abs/2604.04117v1}{arXiv:2604.04117v1}, 2026.

\bibitem[Roffe et~al.(2021)Roffe, Akolkar, George, Linares-Barranco, and
  Benosman]{C04}
Seth Roffe, Himanshu Akolkar, Alan~D. George, Bernab{\'e} Linares-Barranco, and
  Ryad Benosman.
\newblock {Neutron-Induced, Single-Event Effects on Neuromorphic Event-Based
  Vision Sensor: A First Step and Tools to Space Applications}.
\newblock \emph{IEEE Access}, 9:\penalty0 85748--85763, 2021.
\newblock \doi{10.1109/ACCESS.2021.3085136}.
\newblock Author manuscript:
  \href{https://arxiv.org/abs/2102.00112v1}{arXiv:2102.00112v1}.

\bibitem[Rodr{\'{i}}guez-Mart{\'{i}}nez and {P{\'{e}}rez del
  Pulgar}(2025)]{T01}
David Rodr{\'{i}}guez-Mart{\'{i}}nez and C.~J. {P{\'{e}}rez del Pulgar}.
\newblock {Fast Vision in the Dark: A Case for Single-Photon Imaging in
  Planetary Navigation}.
\newblock In \emph{18th Symposium on Advanced Space Technologies in Robotics
  and Automation (ASTRA)}, 2025.
\newblock URL \url{https://arxiv.org/abs/2510.10597v1}.
\newblock Reviewed arXiv:2510.10597v1.

\bibitem[Wang et~al.(2022)Wang, Li, Zhu, Xiang, Huang, and Tian]{T03}
Yixuan Wang, Jianing Li, Lin Zhu, Xijie Xiang, Tiejun Huang, and Yonghong Tian.
\newblock {Learning Stereo Depth Estimation with Bio-Inspired Spike Cameras}.
\newblock In \emph{IEEE International Conference on Multimedia and Expo
  (ICME)}, pages 1--6, 2022.
\newblock \doi{10.1109/ICME52920.2022.9859975}.

\bibitem[Rodr{\'{i}}guez-Mart{\'{i}}nez
  et~al.(2026)Rodr{\'{i}}guez-Mart{\'{i}}nez, van~der Meer, Song, Bera, del
  Pulgar, and Olivares-Mendez]{T02}
David Rodr{\'{i}}guez-Mart{\'{i}}nez, Dave van~der Meer, Junlin Song, Abhishek
  Bera, C.~J.~P{\'{e}}rez del Pulgar, and Miguel~Angel Olivares-Mendez.
\newblock {SPICE-HL3: Single-Photon, Inertial, and Stereo Camera dataset for
  Exploration of High-Latitude Lunar Landscapes}.
\newblock \emph{Scientific Data}, 13:\penalty0 374, 2026.
\newblock \doi{10.1038/s41597-026-06668-8}.

\bibitem[Zhou et~al.(2024)Zhou, Zheng, Lyu, and Wang]{N10}
Jiazhou Zhou, Xu~Zheng, Yuanhuiyi Lyu, and Lin Wang.
\newblock {EventBind: Learning a Unified Representation to Bind Them All for
  Event-based Open-world Understanding}.
\newblock In \emph{European Conference on Computer Vision (ECCV)}, 2024.
\newblock URL
  \url{https://www.ecva.net/papers/eccv_2024/papers_ECCV/html/8931_ECCV_2024_paper.php}.

\bibitem[Liu et~al.(2025)Liu, Li, Zhao, Zhang, Meng, Yu, Ji, and Li]{N09}
Shaoyu Liu, Jianing Li, Guanghui Zhao, Yunjian Zhang, Xin Meng, Fei~Richard Yu,
  Xiangyang Ji, and Ming Li.
\newblock {EventGPT: Event Stream Understanding with Multimodal Large Language
  Models}.
\newblock In \emph{Proceedings of the IEEE/CVF Conference on Computer Vision
  and Pattern Recognition (CVPR)}, pages 29139--29149, 2025.
\newblock URL
  \url{https://openaccess.thecvf.com/content/CVPR2025/html/Liu_EventGPT_Event_Stream_Understanding_with_Multimodal_Large_Language_Models_CVPR_2025_paper.html}.

\bibitem[Zhai et~al.(2026)Zhai, Shi, Guo, Yang, and Wang]{N11}
Jiajun Zhai, Hao Shi, Shangwei Guo, Kailun Yang, and Kaiwei Wang.
\newblock {E-VLA: Event-Augmented Vision-Language-Action Model for Dark and
  Blurred Scenes}.
\newblock \href{https://arxiv.org/abs/2604.04834v2}{arXiv:2604.04834v2}, 2026.

\bibitem[Liu et~al.(2026)Liu, Xu, Zhang, Wang, Chen, Chang, Guo, and
  Kneip]{N12}
Jiaxin Liu, Xun Xu, Zhenhao Zhang, Hanqing Wang, Ruiqi Chen, Shi Chang, Weiyu
  Guo, and Laurent Kneip.
\newblock {Event-VLA: Action-Conditioned Event Fusion for Robust
  Vision-Language-Action Model}.
\newblock \href{https://arxiv.org/abs/2606.29384v1}{arXiv:2606.29384v1}, 2026.

\bibitem[Courtois et~al.(2024)Courtois, Miramond, and Pegatoquet]{N06}
Jonathan Courtois, Beno{\^{i}}t Miramond, and Alain Pegatoquet.
\newblock {Spiking monocular event based 6D pose estimation for space
  application}.
\newblock In \emph{Conference on AI in and for Space (SPAICE)}, pages 394--399,
  2024.
\newblock \doi{10.5281/zenodo.13885633}.

\bibitem[Jones et~al.(2025)Jones, Mees, Sferrazza, Stachowicz, Abbeel, and
  Levine]{G01}
Joshua Jones, Oier Mees, Carmelo Sferrazza, Kyle Stachowicz, Pieter Abbeel, and
  Sergey Levine.
\newblock {Beyond Sight: Finetuning Generalist Robot Policies with
  Heterogeneous Sensors via Language Grounding}.
\newblock \href{https://arxiv.org/abs/2501.04693v3}{arXiv:2501.04693v3}, 2025.

\bibitem[Hafner et~al.(2023)Hafner, Pasukonis, Ba, and Lillicrap]{F20}
Danijar Hafner, Jurgis Pasukonis, Jimmy Ba, and Timothy Lillicrap.
\newblock {Mastering Diverse Domains through World Models}.
\newblock \href{https://arxiv.org/abs/2301.04104v2}{arXiv:2301.04104v2}, 2023.
\newblock Revised 2024-04-17.

\bibitem[Jang et~al.(2025)Jang, Ye, Lin, et~al.]{G12}
Joel Jang, Seonghyeon Ye, Zongyu Lin, et~al.
\newblock {DreamGen: Unlocking Generalization in Robot Learning through Video
  World Models}.
\newblock \href{https://arxiv.org/abs/2505.12705v2}{arXiv:2505.12705v2}, 2025.

\bibitem[Assran et~al.(2025)Assran, Bardes, Fan, et~al.]{T05}
Mido Assran, Adrien Bardes, David Fan, et~al.
\newblock {V-JEPA 2: Self-Supervised Video Models Enable Understanding,
  Prediction and Planning}.
\newblock \href{https://arxiv.org/abs/2506.09985v1}{arXiv:2506.09985v1}, 2025.

\bibitem[Cheang et~al.(2024)Cheang, Chen, Jing, et~al.]{G11}
Chi-Lam Cheang, Guangzeng Chen, Ya~Jing, et~al.
\newblock {GR-2: A Generative Video-Language-Action Model with Web-Scale
  Knowledge for Robot Manipulation}.
\newblock \href{https://arxiv.org/abs/2410.06158v1}{arXiv:2410.06158v1}, 2024.

\bibitem[Orsula et~al.(2026)Orsula, Olivares-Mendez, and Martinez]{F07}
Andrej Orsula, Miguel Olivares-Mendez, and Carol Martinez.
\newblock {Reward-Free Continual Adaptation for Resilient Space Robots}.
\newblock \href{https://arxiv.org/abs/2608.23452v1}{arXiv:2608.23452v1}, 2026.

\bibitem[Huang et~al.(2023)Huang, Wang, Zhang, Li, Wu, and Fei-Fei]{T06}
Wenlong Huang, Chen Wang, Ruohan Zhang, Yunzhu Li, Jiajun Wu, and Li~Fei-Fei.
\newblock {VoxPoser: Composable 3D Value Maps for Robotic Manipulation with
  Language Models}.
\newblock \href{https://arxiv.org/abs/2307.05973v2}{arXiv:2307.05973v2}, 2023.
\newblock Revised 2023-11-02.

\bibitem[Chen et~al.(2026)Chen, Yang, Sun, and Yu]{F06}
HongRun Chen, Li~Yang, Long Sun, and HaoQing Yu.
\newblock {LLM-KGFP: A chain-of-thought-driven knowledge graph planner for
  autonomous rover operations in lunar polar shadowed regions}.
\newblock \emph{Advances in Space Research}, 77\penalty0 (6):\penalty0
  7406--7423, 2026.
\newblock \doi{10.1016/j.asr.2026.01.031}.

\bibitem[Wu et~al.(2026{\natexlab{a}})Wu, Han, Luo, Wang, He, and Wan]{F03}
Aodi Wu, Haodong Han, Xubo Luo, Ruisuo Wang, Shan He, and Xue Wan.
\newblock {SpaceMind: A Modular and Self-Evolving Embodied Vision-Language
  Agent Framework for Autonomous On-orbit Servicing}.
\newblock \href{https://arxiv.org/abs/2604.14399v1}{arXiv:2604.14399v1},
  2026{\natexlab{a}}.

\bibitem[Huang et~al.(2026{\natexlab{b}})Huang, Wei, Cai, and Wang]{F09}
Bingqi Huang, Bingchuan Wei, Yingkai Cai, and Zhaokui Wang.
\newblock {Toward Self-Evolving Lunar Robotic Autonomy Through
  Contract-Governed Skill Registration}.
\newblock \emph{Astronautics}, 1\penalty0 (3):\penalty0 15, 2026{\natexlab{b}}.
\newblock \doi{10.3390/astronautics1030015}.

\bibitem[{NASA Jet Propulsion Laboratory}(2026)]{F12}
{NASA Jet Propulsion Laboratory}.
\newblock {NASA's Perseverance Rover Completes First AI-Planned Drive on Mars}.
\newblock NASA/JPL mission report, 2026.
\newblock URL
  \url{https://www.nasa.gov/missions/mars-2020-perseverance/perseverance-rover/nasas-perseverance-rover-completes-first-ai-planned-drive-on-mars/}.
\newblock Accessed 2026-10-15.

\bibitem[Francis et~al.(2017)Francis, Estlin, Doran, et~al.]{A06}
R.~Francis, T.~Estlin, G.~Doran, et~al.
\newblock {AEGIS autonomous targeting for ChemCam on Mars Science Laboratory:
  Deployment and results of initial science team use}.
\newblock \emph{Science Robotics}, 2\penalty0 (7):\penalty0 eaan4582, 2017.
\newblock \doi{10.1126/scirobotics.aan4582}.

\bibitem[Ren et~al.(2023)Ren, Dixit, Bodrova, et~al.]{A07}
Allen~Z. Ren, Anushri Dixit, Alexandra Bodrova, et~al.
\newblock {Robots That Ask For Help: Uncertainty Alignment for Large Language
  Model Planners}.
\newblock \href{https://arxiv.org/abs/2307.01928v2}{arXiv:2307.01928v2}, 2023.

\bibitem[Carrasco et~al.(2024)Carrasco, Rodriguez-Fernandez, and Linares]{E09}
Alejandro Carrasco, Victor Rodriguez-Fernandez, and Richard Linares.
\newblock {Fine-tuning LLMs for Autonomous Spacecraft Control: A Case Study
  Using Kerbal Space Program}.
\newblock \href{https://arxiv.org/abs/2408.08676v1}{arXiv:2408.08676v1}, 2024.

\bibitem[He et~al.(2026)He, Xie, Cao, et~al.]{T07}
Sida He, Lingxi Xie, Yunning Cao, et~al.
\newblock {Robot Manipulation with GPT-6-Astra: Body Knowledge, Experience
  Reuse, Emergent Skills, and Sim2Real Transfer}.
\newblock \href{https://arxiv.org/abs/2609.31770v1}{arXiv:2609.31770v1}, 2026.

\bibitem[{Galbot Team} et~al.(2026){Galbot Team}, Chen, Cheng, Deng,
  et~al.]{T08}
{Galbot Team}, Xuchuan Chen, Xiaoqian Cheng, Yu~Deng, et~al.
\newblock {Systematically Exploring the Capabilities of GPT-6 Astra as Embodied
  Policies}.
\newblock \href{https://arxiv.org/abs/2609.38537v1}{arXiv:2609.38537v1}, 2026.

\bibitem[Brohan et~al.(2023)Brohan, Brown, Carbajal, et~al.]{F13}
Anthony Brohan, Noah Brown, Justice Carbajal, et~al.
\newblock {RT-2: Vision-Language-Action Models Transfer Web Knowledge to
  Robotic Control}.
\newblock \href{https://arxiv.org/abs/2307.15818v1}{arXiv:2307.15818v1}, 2023.

\bibitem[Chi et~al.(2023)Chi, Feng, Du, Xu, Cousineau, Burchfiel, and
  Song]{F17}
Cheng Chi, Siyuan Feng, Yilun Du, Zhenjia Xu, Eric Cousineau, Benjamin
  Burchfiel, and Shuran Song.
\newblock {Diffusion Policy: Visuomotor Policy Learning via Action Diffusion}.
\newblock \href{https://arxiv.org/abs/2303.04137v1}{arXiv:2303.04137v1}, 2023.

\bibitem[Pertsch et~al.(2025)Pertsch, Stachowicz, Ichter, et~al.]{G08}
Karl Pertsch, Kyle Stachowicz, Brian Ichter, et~al.
\newblock {FAST: Efficient Action Tokenization for Vision-Language-Action
  Models}.
\newblock \href{https://arxiv.org/abs/2501.09747v1}{arXiv:2501.09747v1}, 2025.

\bibitem[Black et~al.(2025)Black, Galliker, and Levine]{U04}
Kevin Black, Manuel~Y. Galliker, and Sergey Levine.
\newblock {Real-Time Execution of Action Chunking Flow Policies}.
\newblock \href{https://arxiv.org/abs/2506.07339v2}{arXiv:2506.07339v2}, 2025.
\newblock Revised 2025-12-05.

\bibitem[Wang et~al.(2026{\natexlab{a}})Wang, Zhang, Wang, Zhang, Yang, and
  Li]{U03}
Jingya Wang, Dehao Zhang, Shuai Wang, Malu Zhang, Yang Yang, and Haizhou Li.
\newblock {SpikingVLA: Asynchronous Spiking Vision-Language-Action Models}.
\newblock \href{https://arxiv.org/abs/2610.09710v1}{arXiv:2610.09710v1},
  2026{\natexlab{a}}.

\bibitem[Yuan et~al.(2026)Yuan, Zhang, Zhou, et~al.]{G04}
Chengbo Yuan, Zicheng Zhang, Mingjie Zhou, et~al.
\newblock {FTP-1: A Generalist Foundation Tactile Policy Across Tactile Sensors
  for Contact-Rich Manipulation}.
\newblock \href{https://arxiv.org/abs/2606.13102v3}{arXiv:2606.13102v3}, 2026.

\bibitem[Xue et~al.(2025)Xue, Ren, Chen, Zhang, Fang, Gu, Xu, and Lu]{U11}
Han Xue, Jieji Ren, Wendi Chen, Gu~Zhang, Yuan Fang, Guoying Gu, Huazhe Xu, and
  Cewu Lu.
\newblock {Reactive Diffusion Policy: Slow-Fast Visual-Tactile Policy Learning
  for Contact-Rich Manipulation}.
\newblock \href{https://arxiv.org/abs/2503.02881v2}{arXiv:2503.02881v2}, 2025.
\newblock Revised 2025-04-01.

\bibitem[Inaba et~al.(2003)Inaba, Oda, and Hayashi]{U12}
Noriyasu Inaba, Mitsushige Oda, and Masato Hayashi.
\newblock {Visual Servoing of Space Robot for Autonomous Satellite Capture}.
\newblock \emph{Transactions of the Japan Society for Aeronautical and Space
  Sciences}, 46\penalty0 (153):\penalty0 173--179, 2003.
\newblock \doi{10.2322/tjsass.46.173}.

\bibitem[Hicsonmez et~al.(2026)Hicsonmez, Sosa, Pineau, Singh, Rathinam,
  Shabayek, and Aouada]{E03}
Samet Hicsonmez, Jose Sosa, Dan Pineau, Inder~Pal Singh, Arunkumar Rathinam,
  Abd El~Rahman Shabayek, and Djamila Aouada.
\newblock {Annotation Free Spacecraft Detection and Segmentation using Vision
  Language Models}.
\newblock \href{https://arxiv.org/abs/2602.04699v1}{arXiv:2602.04699v1}, 2026.

\bibitem[Stewart et~al.(2025{\natexlab{a}})Stewart, Chapin, Leontie, and
  Henshaw]{V17}
Kenneth Stewart, Samantha Chapin, Roxana Leontie, and Carl~Glen Henshaw.
\newblock {Crossing the Sim2Real Gap Between Simulation and Ground Testing to
  Space Deployment of Autonomous Free-flyer Control}.
\newblock \href{https://arxiv.org/abs/2512.03736v1}{arXiv:2512.03736v1},
  2025{\natexlab{a}}.

\bibitem[Taherin et~al.(2026)Taherin, Lin, Akbari, Akbari, Zhao, Chen, Kaeli,
  and Wang]{T14}
Amir Taherin, Juyi Lin, Arash Akbari, Arman Akbari, Pu~Zhao, Weiwei Chen, David
  Kaeli, and Yanzhi Wang.
\newblock Cross-platform scaling of vision-language-action models from edge to
  cloud {GPUs}, 2026.
\newblock URL \url{https://arxiv.org/abs/2509.11480v2}.
\newblock Reviewed version 2, revised 26 January 2026; first submitted 2025.

\bibitem[Wang et~al.(2026{\natexlab{b}})Wang, Xiong, Wang, and Chen]{T15}
Hongyu Wang, Chuyan Xiong, Ruiping Wang, and Xilin Chen.
\newblock {BitVLA}: 1-bit vision-language-action models for robotics
  manipulation, 2026{\natexlab{b}}.
\newblock URL \url{https://arxiv.org/abs/2506.07530v2}.
\newblock Reviewed version 2, revised 1 March 2026; first submitted 2025.

\bibitem[Kim et~al.(2025)Kim, Finn, and Liang]{U05}
Moo~Jin Kim, Chelsea Finn, and Percy Liang.
\newblock {Fine-Tuning Vision-Language-Action Models: Optimizing Speed and
  Success}.
\newblock \href{https://arxiv.org/abs/2502.19645v2}{arXiv:2502.19645v2}, 2025.
\newblock Revised 2025-04-28.

\bibitem[Zhang et~al.(2025)Zhang, Zhou, Shao, Su, and Sun]{N21}
Liwen Zhang, Dong Zhou, Shibo Shao, Zihao Su, and Guanghui Sun.
\newblock {Multimodal Spiking Neural Network for Space Robotic Manipulation}.
\newblock \href{https://arxiv.org/abs/2508.07287v1}{arXiv:2508.07287v1}, 2025.

\bibitem[Song et~al.(2026)Song, Nie, Teng, et~al.]{U01}
Ruiqi Song, Dujun Nie, Siyu Teng, et~al.
\newblock {SpikeVLA: Vision-Language-Action Models with Spiking Neural
  Networks}.
\newblock \href{https://arxiv.org/abs/2606.27807v1}{arXiv:2606.27807v1}, 2026.

\bibitem[Wang et~al.(2026{\natexlab{c}})Wang, Zhang, Liu, et~al.]{U02}
Shuai Wang, Malu Zhang, Mingquan Liu, et~al.
\newblock {Spike-driven Vision-Language-Action Model}.
\newblock \href{https://arxiv.org/abs/2609.39514v1}{arXiv:2609.39514v1},
  2026{\natexlab{c}}.

\bibitem[Clark et~al.(2026)Clark, Wang, Gao, Hong, Choi, Cutkosky, Hou, and
  Song]{G14}
Jaden Clark, Changhao Wang, Yihuai Gao, Seongheon Hong, Hojung Choi, Mark
  Cutkosky, Yifan Hou, and Shuran Song.
\newblock {Multisensory Continual Learning: Adapting Pretrained Visuomotor
  Policies to Force}.
\newblock \href{https://arxiv.org/abs/2606.30988v4}{arXiv:2606.30988v4}, 2026.

\bibitem[Yik et~al.(2025)Yik, {Van den Berghe}, {den Blanken}, et~al.]{U06}
Jason Yik, Korneel {Van den Berghe}, Douwe {den Blanken}, et~al.
\newblock {The neurobench framework for benchmarking neuromorphic computing
  algorithms and systems}.
\newblock \emph{Nature Communications}, 16:\penalty0 1545, 2025.
\newblock \doi{10.1038/s41467-025-56739-4}.

\bibitem[Paredes-Vall{\'{e}}s et~al.(2024)Paredes-Vall{\'{e}}s, Hagenaars,
  Dupeyroux, Stroobants, Xu, and de~Croon]{N08}
Federico Paredes-Vall{\'{e}}s, Jesse Hagenaars, Julien Dupeyroux, Stein
  Stroobants, Yingfu Xu, and Guido de~Croon.
\newblock {Fully neuromorphic vision and control for autonomous drone flight}.
\newblock \emph{Science Robotics}, 9:\penalty0 eadi0591, 2024.
\newblock \doi{10.1126/scirobotics.adi0591}.
\newblock URL \url{https://arxiv.org/abs/2303.08778v1}.
\newblock Energy benchmark inspected in preprint arXiv:2303.08778v1, Table 2.

\bibitem[Wang et~al.(2024{\natexlab{b}})Wang, Qiu, Xu, Wang, Li, Zhang, Liu,
  and Li]{P09}
Meiqi Wang, Han Qiu, Longnv Xu, Di~Wang, Yuanjie Li, Tianwei Zhang, Jun Liu,
  and Hewu Li.
\newblock {A Case for Application-Aware Space Radiation Tolerance in Orbital
  Computing}.
\newblock \href{https://arxiv.org/abs/2407.11853v1}{arXiv:2407.11853v1},
  2024{\natexlab{b}}.

\bibitem[Xue et~al.(2023)Xue, Liu, Wang, Yang, Luo, Zhang, Li, and Li]{N17}
Xinghua Xue, Cheng Liu, Ying Wang, Bing Yang, Tao Luo, Lei Zhang, Huawei Li,
  and Xiaowei Li.
\newblock {Soft Error Reliability Analysis of Vision Transformers}.
\newblock \href{https://arxiv.org/abs/2302.10468v3}{arXiv:2302.10468v3}, 2023.
\newblock Revised 2024-04-26.

\bibitem[Frias-Dominguez et~al.(2026)Frias-Dominguez, Badia, Leon, Amor-Martin,
  and Belloch]{N22}
Lester Frias-Dominguez, Jose~M. Badia, German Leon, Adrian Amor-Martin, and
  Jose~A. Belloch.
\newblock {Dependability analysis and hardening of vision transformers against
  soft errors}.
\newblock \emph{The Journal of Supercomputing}, 82:\penalty0 331, 2026.
\newblock \doi{10.1007/s11227-026-08373-0}.

\bibitem[Mees et~al.(2021)Mees, Hermann, Rosete-Beas, and Burgard]{A10}
Oier Mees, Lukas Hermann, Erick Rosete-Beas, and Wolfram Burgard.
\newblock {CALVIN: A Benchmark for Language-Conditioned Policy Learning for
  Long-Horizon Robot Manipulation Tasks}.
\newblock \href{https://arxiv.org/abs/2112.03227v4}{arXiv:2112.03227v4}, 2021.
\newblock Revised 2022-07-13.

\bibitem[Liu et~al.(2023)Liu, Zhu, Gao, Feng, Liu, Zhu, and Stone]{A11}
Bo~Liu, Yifeng Zhu, Chongkai Gao, Yihao Feng, Qiang Liu, Yuke Zhu, and Peter
  Stone.
\newblock {LIBERO: Benchmarking Knowledge Transfer for Lifelong Robot
  Learning}.
\newblock \href{https://arxiv.org/abs/2306.03310v2}{arXiv:2306.03310v2}, 2023.

\bibitem[Li et~al.(2024)Li, Hsu, Gu, et~al.]{A12}
Xuanlin Li, Kyle Hsu, Jiayuan Gu, et~al.
\newblock {Evaluating Real-World Robot Manipulation Policies in Simulation}.
\newblock \href{https://arxiv.org/abs/2405.05941v1}{arXiv:2405.05941v1}, 2024.

\bibitem[Wu et~al.(2026{\natexlab{b}})Wu, Zuo, Zhao, Luo, Wang, and Wan]{P08}
Aodi Wu, Jianhong Zuo, Zeyuan Zhao, Xubo Luo, Ruisuo Wang, and Xue Wan.
\newblock {SpaceSense-Bench: A Large-Scale Multi-Modal Benchmark for Spacecraft
  Perception and Pose Estimation}.
\newblock \href{https://arxiv.org/abs/2603.09320v1}{arXiv:2603.09320v1},
  2026{\natexlab{b}}.

\bibitem[Orsula et~al.(2025)Orsula, Geist, Olivares-Mendez, and Martinez]{V15}
Andrej Orsula, Matthieu Geist, Miguel Olivares-Mendez, and Carol Martinez.
\newblock {Space Robotics Bench: Robot Learning Beyond Earth}.
\newblock \href{https://arxiv.org/abs/2509.23328v1}{arXiv:2509.23328v1}, 2025.

\bibitem[Kenneally et~al.(2020)Kenneally, Piggott, and Schaub]{V10}
Patrick~W. Kenneally, Scott Piggott, and Hanspeter Schaub.
\newblock {Basilisk: A Flexible, Scalable and Modular Astrodynamics Simulation
  Framework}.
\newblock \emph{Journal of Aerospace Information Systems}, 17\penalty0
  (9):\penalty0 496--507, 2020.
\newblock \doi{10.2514/1.I010762}.

\bibitem[Rebecq et~al.(2018)Rebecq, Gehrig, and Scaramuzza]{N23}
Henri Rebecq, Daniel Gehrig, and Davide Scaramuzza.
\newblock {ESIM: An Open Event Camera Simulator}.
\newblock In \emph{Proceedings of the 2nd Conference on Robot Learning},
  volume~87 of \emph{Proceedings of Machine Learning Research}, pages 969--982,
  2018.
\newblock URL \url{https://proceedings.mlr.press/v87/rebecq18a.html}.

\bibitem[Hu et~al.(2021)Hu, Liu, and Delbruck]{N24}
Yuhuang Hu, Shih-Chii Liu, and Tobi Delbruck.
\newblock {v2e: From Video Frames to Realistic DVS Events}.
\newblock In \emph{IEEE/CVF Conference on Computer Vision and Pattern
  Recognition Workshops}, 2021.
\newblock URL \url{https://arxiv.org/abs/2006.07722v2}.
\newblock Reviewed preprint arXiv:2006.07722v2; first posted 2020.

\bibitem[Dvo{\v{r}}{\'{a}}k et~al.(2025)Dvo{\v{r}}{\'{a}}k, M{\"{a}}rtens, and
  Izzo]{N18}
Ond{\v{r}}ej Dvo{\v{r}}{\'{a}}k, Marcus M{\"{a}}rtens, and Dario Izzo.
\newblock {Event-based Lunar OPtical flow Egomotion estimation (ELOPE)
  dataset}.
\newblock Zenodo, version 1, 2025.
\newblock URL \url{https://zenodo.org/records/15421707}.

\bibitem[Sikorski et~al.(2021)Sikorski, Izzo, and Meoni]{N19}
Olaf Sikorski, Dario Izzo, and Gabriele Meoni.
\newblock {Event-Based Spacecraft Landing Using Time-To-Contact}.
\newblock In \emph{Proceedings of the IEEE/CVF Conference on Computer Vision
  and Pattern Recognition Workshops}, pages 1941--1950, 2021.
\newblock URL
  \url{https://openaccess.thecvf.com/content/CVPR2021W/AI4Space/html/Sikorski_Event-Based_Spacecraft_Landing_Using_Time-To-Contact_CVPRW_2021_paper.html}.

\bibitem[Xiao et~al.(2026)Xiao, Tian, Yin, He, Wang, Liu, and Xia]{F08}
Xin-Yu Xiao, Ye~Tian, Erwei Yin, Zhixian He, Shiqi Wang, Yalei Liu, and
  Qianchen Xia.
\newblock {Lunar-Bench: Towards Evaluating Task-Oriented Reasoning of LLMs in
  Lunar Exploration Scenarios}.
\newblock In \emph{Findings of the Association for Computational Linguistics:
  ACL 2026}, pages 1668--1705, 2026.
\newblock \doi{10.18653/v1/2026.findings-acl.83}.

\bibitem[Li et~al.(2021)Li, Wei, Liang, Ren, Tang, and Li]{P06}
Haiquan Li, Qingqing Wei, Jianxun Liang, Weiyan Ren, Zixin Tang, and Delun Li.
\newblock {Estimation of Vibration Characteristics of a Space Manipulator From
  Air Bearing Supported Test Data}.
\newblock \emph{Frontiers in Robotics and AI}, 8:\penalty0 641165, 2021.
\newblock \doi{10.3389/frobt.2021.641165}.

\bibitem[Elhardt et~al.(2023)Elhardt, Boumann, De~Stefano, et~al.]{P04}
Ferdinand Elhardt, Roland Boumann, Marco De~Stefano, et~al.
\newblock {The Motion Suspension System -- MSS: A Cable-Driven System for
  On-Ground Tests of Space Robots}.
\newblock In \emph{Advances in Mechanism and Machine Science: Proceedings of
  the 16th IFToMM World Congress}, volume 148 of \emph{Mechanisms and Machine
  Science}, pages 379--388. Springer, 2023.
\newblock \doi{10.1007/978-3-031-45770-8_38}.

\bibitem[Elhardt et~al.(2024)Elhardt, de~Stefano, Schedl, et~al.]{P05}
Ferdinand Elhardt, Marco de~Stefano, Manfred Schedl, et~al.
\newblock {The Motion Suspension System for On-Ground Tests of Space Robots:
  Demonstration and Results}.
\newblock In \emph{IEEE Aerospace Conference}, 2024.
\newblock \doi{10.1109/AERO58975.2024.10521208}.

\bibitem[{DLR}(n.d.)]{V06}
{DLR}.
\newblock {European Proximity Operations Simulator (EPOS 2.0)}.
\newblock DLR research infrastructure, n.d.
\newblock URL
  \url{https://www.dlr.de/en/research-and-transfer/research-infrastructure/european-proximity-operations-simulator-epos}.
\newblock Accessed 2026-10-15.

\bibitem[Artigas et~al.(2015)Artigas, De~Stefano, Rackl, et~al.]{V08}
Jordi Artigas, Marco De~Stefano, Wolfgang Rackl, et~al.
\newblock {The OOS-SIM: An On-ground Simulation Facility For On-Orbit Servicing
  Robotic Operations}.
\newblock In \emph{IEEE International Conference on Robotics and Automation
  (ICRA)}, pages 2854--2860, 2015.
\newblock URL \url{https://elib.dlr.de/97179/1/Artigas_OOSSIM_icra15.pdf}.

\bibitem[Zebenay et~al.(2013)Zebenay, Boge, Krenn, and Choukroun]{V07}
Melak Zebenay, Toralf Boge, Rainer Krenn, and Daniel Choukroun.
\newblock {Analytical and experimental stability investigation of a
  hardware-in-the-loop satellite docking simulator}.
\newblock \href{https://arxiv.org/abs/1309.3512v1}{arXiv:1309.3512v1}, 2013.

\bibitem[Stewart et~al.(2025{\natexlab{b}})Stewart, Leontie, Chapin, Hays,
  Shrestha, and Henshaw]{V18}
Kenneth Stewart, Roxana Leontie, Samantha Chapin, Joe Hays, Sumit~Bam Shrestha,
  and Carl~Glen Henshaw.
\newblock {Autonomous Reinforcement Learning Robot Control with Intel's Loihi 2
  Neuromorphic Hardware}.
\newblock \href{https://arxiv.org/abs/2512.03911v1}{arXiv:2512.03911v1},
  2025{\natexlab{b}}.

\bibitem[Scrofano et~al.(2024)Scrofano, Davis, and Taggart]{P12}
Ronald Scrofano, Scott Davis, and Jennifer Taggart.
\newblock {Radiation Test Performance of the Intel Loihi Neuromorphic
  Processor}.
\newblock In \emph{IEEE Space Computing Conference (SCC)}, 2024.
\newblock \doi{10.1109/SCC61854.2024.00016}.

\bibitem[{NASA}(2024)]{U13}
{NASA}.
\newblock {Standard for Models and Simulations}.
\newblock Technical Report NASA-STD-7009B, {NASA}, 2024.
\newblock URL
  \url{https://standards.nasa.gov/sites/default/files/standards/NASA/B/1/NASA-STD-7009B-Final-3-5-2024.pdf}.
\newblock Approved 2024-03-05.

\bibitem[Orsula et~al.(2022)Orsula, B{\o}gh, Olivares-Mendez, and
  Martinez]{V16}
Andrej Orsula, Simon B{\o}gh, Miguel Olivares-Mendez, and Carol Martinez.
\newblock {Learning to Grasp on the Moon from 3D Octree Observations with Deep
  Reinforcement Learning}.
\newblock In \emph{IEEE/RSJ International Conference on Intelligent Robots and
  Systems (IROS)}, pages 4112--4119, 2022.
\newblock \doi{10.1109/IROS47612.2022.9981661}.

\bibitem[McHarg et~al.(2022)McHarg, Balthazor, McReynolds, et~al.]{T04}
Matthew~G. McHarg, Richard~L. Balthazor, Brian~J. McReynolds, et~al.
\newblock {Falcon Neuro: an event-based sensor on the International Space
  Station}.
\newblock \emph{Optical Engineering}, 61\penalty0 (8):\penalty0 085105, 2022.
\newblock \doi{10.1117/1.OE.61.8.085105}.

\bibitem[Djebko et~al.(2026)Djebko, Baumann, Dilger, Puppe, and
  Montenegro]{C01}
Kirill Djebko, Tom Baumann, Erik Dilger, Frank Puppe, and Sergio Montenegro.
\newblock {LeLaR: The First In-Orbit Demonstration of an AI-Based Satellite
  Attitude Controller}.
\newblock \emph{IEEE Access}, 14:\penalty0 49348--49367, 2026.
\newblock \doi{10.1109/ACCESS.2026.3678816}.
\newblock Reviewed author version:
  \href{https://arxiv.org/abs/2512.19576v5}{arXiv:2512.19576v5}.

\bibitem[Agarwal et~al.(2021)Agarwal, Schwarzer, Castro, Courville, and
  Bellemare]{U14}
Rishabh Agarwal, Max Schwarzer, Pablo~Samuel Castro, Aaron~C. Courville, and
  Marc Bellemare.
\newblock {Deep Reinforcement Learning at the Edge of the Statistical
  Precipice}.
\newblock In \emph{Advances in Neural Information Processing Systems},
  volume~34, pages 29304--29320, 2021.
\newblock URL
  \url{https://papers.neurips.cc/paper/2021/hash/f514cec81cb148559cf475e7426eed5e-Abstract.html}.
\newblock Reviewed arXiv:2108.13264v4, revised 2022-01-05.

\end{thebibliography}
\end{document}